\documentclass[referee,pdflatex,sn-mathphys-num]{sn-jnl}

\usepackage{graphicx}%
\usepackage{multirow}%
\usepackage{amsmath,amssymb,amsfonts}%
\usepackage{amsthm}%
\usepackage{mathrsfs}%
\usepackage{xcolor}%
\usepackage{textcomp}%
\usepackage{booktabs}%
\usepackage{tabularx,array}%

\newcolumntype{Y}{>{\centering\arraybackslash}X}

\begin{document}

\title[Self-Supervised Weighted Image Guided qMRI Super-Resolution]{Self-Supervised Weighted Image Guided Quantitative MRI Super-Resolution}

\author*[1]{\fnm{Alireza} \sur{Samadifardheris}}\email{a.samadifardheris@erasmusmc.nl}

\author[1]{\fnm{Dirk H.J.} \sur{Poot}}\email{d.poot@erasmusmc.nl}

\author[2]{\fnm{Florian} \sur{Wiesinger}}\email{florian.wiesinger@gehealthcare.com}

\author[1]{\fnm{Stefan} \sur{Klein}}\email{s.klein@erasmusmc.nl}

\author[1,3]{\fnm{Juan A.} \sur{Hernandez-Tamames}}\email{j.hernandeztamames@erasmusmc.nl}

\affil*[1]{\orgdiv{Department of Radiology and Nuclear Medicine}, \orgname{Erasmus MC}, \orgaddress{\city{Rotterdam}, \country{The Netherlands}}}

\affil[2]{\orgname{GE HealthCare}, \orgaddress{\city{Munich}, \country{Germany}}}

\affil[3]{\orgdiv{Department of Imaging Physics}, \orgname{Delft University of Technology}, \orgaddress{\city{Delft}, \country{The Netherlands}}}

\abstract{\textbf{Object:} To present and evaluate Self-supervised Weighted Image Guided quantitative MRI Super-Resolution (SWIG qMRI SR), a physics-informed framework recovering high-resolution (HR) qMRI from a rapid low-resolution (LR) acquisition guided by routine weighted images (wMRI), without HR training targets.
\textbf{Materials and Methods:} A CNN matches acquired wMRI to images synthesized from predicted maps through forward signal models, while anchoring those maps to the acquired LR qMRI. Training and ablation used synthetic data (n\,=\,27); cross-sequence generalizability was tested in three volunteers scanned with silent 3D MuPa-ZTE at 1 and 5\,min, with clinical T1w/T2w guides.
\textbf{Results:} In synthetic data, dual-guide SR reduced the T1 high-frequency error norm 45.4\% below baseline; removing the LR qMRI anchor raised gray-matter T1 error from 63 to 110\,ms. In vivo, super-resolved 1-min maps synthesized T1w images (SSIM 0.93, PSNR 27.3\,dB) surpassing synthesis from 5-min maps (0.83), and improved synthesis of T2-FLAIR, never used as a guide (SSIM 0.69 to 0.75).
\textbf{Discussion:} HR detail was recovered without HR supervision, and the network transferred across a different qMRI sequence. Because the guides are already acquired, adding a 1-min qMRI scan to a standard exam offers a practical route to routine clinical qMRI integration.}


\maketitle

\section{Introduction}\label{sec1}

Magnetic resonance imaging (MRI) offers both conventional contrast-weighted images (wMRI), which are widely used in clinical practice for their diagnostic value, and quantitative MRI (qMRI) techniques, such as Proton Density (PD), T1 and T2 relaxometry, which provide objective tissue-specific measurements. While wMRI scans rely on subjective visual interpretation and are less reproducible, qMRI provides objective and consistent biomarkers that are less sensitive to scanner hardware or operator variability~\cite{bib_b1,bib_b2}.

Traditionally, qMRI maps are obtained by fitting a signal model across a series of wMRI acquired with varying imaging parameters to provide sufficient contrast diversity to allow estimation of all parameters of the model. Acquiring such a series---particularly at high resolution (HR)---is time-consuming and typically not possible in clinical settings with patients. To accelerate qMRI acquisition, deep learning (DL) techniques have emerged as powerful tools, but they differ fundamentally in their approach to the acquisition-resolution trade-off. One class of methods directly estimates qMRI parameters from undersampled k-space data. For example, Liu et al.~\cite{bib_b3} proposed a physics-informed mapping to incorporate physics-based signal models into the network architecture to estimate parameters from undersampled multi-echo wMRI. While this approach accelerates acquisition by reducing k-space sampling requirements, it faces inherent SNR limitations: acquiring sufficient signal from small voxels for HR mapping still requires longer scan times to maintain adequate signal-to-noise ratio. Although DL can partially compensate through learned priors, overly strong priors risk suppressing subtle pathological features, introducing a fundamental trade-off between spatial resolution, acquisition time, SNR, and quantitative reliability.

An alternative strategy involves rapidly acquiring low resolution (LR) qMRI and subsequently applying super-resolution (SR) techniques to enhance the image quality. The ``supervised SR'' approach of Chaudhari et al.~\cite{bib_b4}, developed for wMRI, can be applied to qMRI SR but faces a fundamental limitation: the requirement for HR ground truth during training. For qMRI specifically, obtaining HR parametric maps at the scale and diversity needed for supervised training is even more challenging. Most available qMRI datasets are limited to healthy volunteers, making it difficult for supervised models to generalize to pathological cases where tissue properties may differ substantially from training data.

To address this training data bottleneck, recent self-supervised strategies for direct ``wMRI-to-qMRI'' mapping aim to learn HR qMRI directly from routine wMRI without requiring HR qMRI for training. For instance, Qiu et al.~\cite{bib_b5} proposed a framework where predicted qMRI maps are passed through a known forward signal model to synthesize corresponding wMRI, with the synthesized images compared to acquired wMRI as supervision. This eliminates the need for HR qMRI training labels. However, in subsequent work~\cite{bib_b6}, the same authors relied on three conventional wMRI as inputs while supervising with ground-truth qMRI. Similarly, Moya-S\'aez et al.~\cite{bib_b7} estimate HR qMRI from routine clinical images but still require HR qMRI during training for supervision, restricting applicability in settings where such data is scarce. Furthermore, methods relying solely on routine clinical wMRI for self-supervision are constrained by the number and diversity of available contrasts, which in standard clinical protocols is often insufficient to uniquely determine all parameters of the signal model, thereby introducing a strong dependency on a learned prior.

A parallel line of research has explored using side information---typically high-resolution contrast-weighted images from complementary sequences---to improve reconstruction quality in accelerated MRI. Methods such as TGVN~\cite{bib_b8} demonstrate that incorporating auxiliary contrast images can disambiguate ill-posed inverse problems and improve spatial resolution. However, these approaches operate on contrast images rather than quantitative parametric maps, addressing the fundamentally different problem of k-space undersampling during acquisition. While conceptually related through their use of complementary imaging data, side information methods for contrast reconstruction cannot directly address qMRI super-resolution: contrast images are sequence-dependent and lack the quantitative, scanner-independent tissue properties that define relaxometry parameters. Our work bridges these domains by formulating qMRI SR as a physics-informed problem that leverages routine clinical contrast images as spatial priors while preserving the quantitative nature of the parametric maps.

In summary, two divergent paradigms have emerged: (1) bypassing qMRI
acquisition entirely by estimating HR qMRI directly from conventional wMRI
using DL, and (2) acquiring LR qMRI followed by supervised SR using HR qMRI
ground truth for training. The former requires no quantitative acquisition and
therefore adds no scan time, but at deployment it produces patient-specific
parametric maps entirely from priors learned on a training population, with no
acquired quantitative measurement of that subject against which they can be
checked. The latter preserves the quantitative measurement within the clinical
workflow and enables rigorous patient-specific validation by comparison with the
acquired LR qMRI, but it is constrained by the practical difficulty of
acquiring sufficient HR qMRI training data, especially in patient populations.
This raises critical questions on how best to balance acquisition efficiency
with data fidelity while enabling integration of qMRI into clinical practice:
Can qMRI SR be guided using only data already available in routine clinical
protocols---thereby minimizing additional acquisition burden while leveraging
highly representative data---to facilitate broader clinical adoption and
integration of fast qMRI into standard imaging workflows for objective,
rigorous validation?

\begin{table*}[!h]
\centering
\renewcommand{\arraystretch}{1.4}
\caption{Comparison of SWIG qMRI SR with representative prior
methods across four key design properties. \checkmark~=~satisfied;
\texttimes~=~not satisfied; --~=~not applicable to this method's
problem domain or not evaluated; (1) No HR qMRI supervision target required;
(2) SR guided by routine clinical wMRI; (3) Acquired LR qMRI preserved
as quantitative anchor; (4) Cross-sequence generalizability.}
\label{tab:method-comparison}
\begin{tabularx}{\textwidth}{>{\hsize=1.4\hsize}X
                              >{\hsize=0.65\hsize}Y
                              >{\hsize=0.65\hsize}Y
                              >{\hsize=0.65\hsize}Y
                              >{\hsize=0.65\hsize}Y}
\toprule
\textbf{Method} &
\textbf{(1) No HR qMRI} &
\textbf{(2) wMRI guided} &
\textbf{(3) LR anchor} &
\textbf{(4) Cross-sequence} \\
\midrule
Physics-informed mapping~\cite{bib_b3}
  & \checkmark & \texttimes & \texttimes & \texttimes \\

Supervised SR~\cite{bib_b4}
  & \texttimes & \texttimes & \checkmark & -- \\

Self-Supervised wMRI-to-qMRI~\cite{bib_b5}
  & \checkmark & \checkmark & \texttimes & \texttimes \\

Supervised wMRI-to-qMRI~\cite{bib_b6,bib_b7}
  & \texttimes & \checkmark & \texttimes & \texttimes \\

Side-info recon.~\cite{bib_b8}
  & -- & \checkmark & -- & -- \\

\textbf{SWIG qMRI SR (ours)}
  & \checkmark & \checkmark & \checkmark & \checkmark \\
\bottomrule
\end{tabularx}
\end{table*}

\subsection{Proposed Framework}\label{subsec1}

We propose a framework termed \textbf{S}elf-Supervised \textbf{W}eighted \textbf{I}mage \textbf{G}uided \textbf{qMRI} \textbf{S}uper-\textbf{R}esolution (\textbf{SWIG qMRI SR}) that enhances the resolution of LR qMRI using HR wMRI scans as guidance. While the use of HR wMRI as guidance resembles the methods by Atal\i k et al.~\cite{bib_b8} that leverage side information for MRI reconstruction, their approach reconstructs contrast images from undersampled k-space---operating within wMRI domain. In contrast, SWIG qMRI SR addresses qMRI SR using wMRI as guides, requiring physics-based forward models to bridge these distinct domains while preserving absolute quantitative tissue values.
Our approach differs from existing qMRI SR methods in that it does not require HR parametric maps as training targets, nor the acquisition of additional wMRI beyond standard clinical protocols. Instead, we employ a physics-informed, self-supervised training strategy that minimizes the discrepancy between acquired HR wMRI and corresponding images synthesized from predicted HR qMRI maps via forward signal models and keeping predicted HR qMRI consistent with LR qMRI inputs. This design utilizes HR wMRI data already present in routine clinical imaging workflows.

By operating on qMRI maps rather than raw contrast series of qMRI acquisition, our framework decouples from protocol-specific acquisition choices in the quantitative mapping process itself. This design choice is hypothesized to enable cross-qMRI sequence (cross-sequence) generalization: since the method operates on scanner-independent tissue relaxometry parameters rather than sequence-dependent signal intensities, it may allow application to qMRI data from different acquisition protocols beyond those used during training.
SWIG qMRI SR combines four key design properties: (1) no requirement for HR parametric maps during training, (2) SR guidance from routine clinical wMRI, (3) preservation of acquired LR qMRI measurements for quantitative validation and traceability, and (4) cross-sequence generalizability by operating on protocol-independent tissue relaxometry parameters rather than sequence-specific signal intensities. This combination distinguishes the proposed approach from methods that either bypass qMRI acquisition entirely or require HR qMRI training data.

No prior method simultaneously satisfies all four properties. Physics-informed mapping methods~\cite{bib_b3} incorporate acquisition-side constraints but do not leverage routinely available clinical wMRI guidance, violating (2), and remain protocol-specific, violating (4). Supervised SR methods~\cite{bib_b4} preserve quantitative traceability through the LR qMRI anchor but require HR qMRI training data, violating (1). Self-supervised wMRI-to-qMRI approaches~\cite{bib_b5} satisfy property (1) if sufficient wMRI are available, but discard acquired LR qMRI measurements as a quantitative reference, violating (3), and operate on sequence-dependent signal intensities, violating (4). Supervised wMRI-to-qMRI approaches~\cite{bib_b6,bib_b7} similarly leverage routine clinical wMRI, satisfying (2), but require HR qMRI as a training target, violating (1); they additionally discard the acquired LR qMRI as a quantitative anchor, violating (3), and operate on sequence-dependent signal intensities, violating (4). Side-information reconstruction approaches~\cite{bib_b8} leverage auxiliary clinical imaging guidance but are not formulated for qMRI SR. The contribution of SWIG qMRI SR is the identification and closure of this gap, as summarized in Table~\ref{tab:method-comparison}.

\section{Materials and Methods}\label{sec2}

\subsection{Notation}\label{subsec2}
Let $Q$ represent a set of qMRI maps (PD, T1, T2), with subscripts $l$ and $h$ denoting LR and HR versions, respectively. Parametric maps $\hat{Q}$ are estimated from a set of contrast images $C = \{ C^i\quad \forall i \}$ by a signal-fitting process $\mathcal{S}$

\begin{equation}
\hat{Q} = \mathcal{S}(C) = \arg\min_Q \sum_i \| C^i - \mathcal{M}^i(Q) \|^2,
\label{Signalfitting}
\end{equation}
where $\mathcal{M}^i$ is the forward model that models the $i$-th wMRI:
\begin{equation}
\hat{C}^i = \mathcal{M}^i(\hat{Q}).
\label{phmodel}
\end{equation}

Operator \( \Phi \) simulates $k$-space undersampling to generate LR from HR contrast images
\begin{equation}
C^i_l = \Phi^i(C^i_h) = \mathcal{F}^{-1} ( (\mathcal{F}(C^i_h) + \epsilon) \circ B^i ),
\label{partialsampling}
\end{equation}
where \( B^i \) is a binary $k$-space mask that is multiplied element-wise (denoted by \( \circ \)), \( \mathcal{F} \) and \( \mathcal{F}^{-1} \) denote the Fourier and inverse Fourier transforms respectively and \( \epsilon \sim \mathcal{N}_c(0, \sigma^2) \) is a realization of zero mean complex Gaussian noise, with variance $\sigma^2$ independently drawn for each voxel.

For each guide $W^j$ the forward model is
\begin{equation}
    W^j = \mathcal{M}_g^j(Q_h),
    \label{phmodel1}
\end{equation}
where the index $j \in \{1, \ldots, N_g\}$ iterates over the set of guide images, e.g., $j=1$ for T1-weighted (T1w), $j=2$ for T2-weighted (T2w). In summary, $C$ denotes contrast images acquired as part of the qMRI protocol and used for parameter fitting, while $W$ denotes independently acquired routine clinical weighted images serving exclusively as HR structural guides.

\subsection{SWIG qMRI SR Framework}\label{subsec3}
The framework estimates HR qMRI, $\hat{Q}_h$, from $Q_l$ using guidance from conventional HR wMRI, $W$. In the derivation below, we assume all images are resampled to the same voxel grid. We define SR as the recovery of high-spatial-frequency information that is absent in $Q_l$ but present in $W$ and representable at the shared voxel grid---thus, SR refers to enhancement of spatial frequency content rather than voxel count increase.

To formalize the relationship between $Q_l$, \(W\), and the desired HR estimate, $\hat{Q}_h$, we start from Bayes' rule

\begin{equation}
    p(\hat{Q}_h \mid C_l,W)
    = \frac{p(C_l, W \mid \hat{Q}_h) \cdot p(\hat{Q}_h)}{p(C_l, W)}.
    \label{bayes_rule}
\end{equation}
This posterior probability reflects how likely a candidate HR map $\hat{Q}_h$ is, given the observed data $(C_l, W)$. The denominator \( p(C_l, W) \) is a normalizing constant (independent of $\hat{Q}_h$) and can be ignored when formulating an optimization objective. Assuming a non-informative (uniform) prior on $\hat{Q}_h$ and conditional independence of $C_l$ and $W$ given $\hat{Q}_h$, the posterior simplifies to

\begin{equation}
  p(\hat{Q}_h \mid C_l, W)  \propto
  p(C_l \mid \hat{Q}_h) \cdot p(W \mid \hat{Q}_h).
\label{eq:likelihood}
\end{equation}
To express these likelihood terms explicitly, we invoke the forward models $\mathcal{M}^i$ and $\mathcal{M}_g^j$ (\eqref{phmodel}, \eqref{phmodel1}) and downsampling operator $\Phi^i$ (\eqref{partialsampling}) to synthesize predicted observations $\hat{C}_l^i$ and $\hat{W}^j$ from candidate HR maps $\hat{Q}_h$. Maximum a posteriori (MAP) inference gives

\begin{equation}
\hat{Q}_h = \arg\max_{\hat{Q}_h} \big[ p(C_l \mid \hat{C}_l)    \cdot p(W \mid \hat{W})  \big] ,
\label{1}
\end{equation}
which is equivalent to minimizing the negative log-likelihood

\begin{align}
L =  - \log p(C_l \mid \hat{C}_l) -\log p(W \mid \hat{W}) = L_l + L_h.
\label{eq:neg_log_likelihood}
\end{align}

The LR data consistency term $L_l$ enforces absolute intensity matching between acquired and predicted contrast images. The contrast images $C_l^i$ (from qMRI acquisition) and $\hat{C}_l^i$ (synthesized from $\hat{Q}_h$ via forward model $\mathcal{M}^i$ and subsequently downsampled via $\Phi^i$) represent the same physical measurement at different processing stages---they must match in absolute signal values to preserve quantitative accuracy. Any metric that permits arbitrary scaling of intensities would allow $\hat{Q}_h$ values that satisfy the metric while destroying the quantitative nature of the parametric maps. Assuming Gaussian distribution for $p(C_l \mid \hat{C}_l)$, we define

\begin{equation}
\begin{aligned}
    -\log p(C_l \mid \hat{C}_l) = L_{\text{l}}
    &=   \sum_{i} ( \text{MSE}(C_l^i, \hat{C}_l^i)/\sigma^2 ),
\end{aligned}
\label{hl}
\end{equation}

where MSE enforces voxel-wise intensity agreement necessary for maintaining absolute quantitative values.

\textbf{Handling Intensity Variations in Guide Images}: MR images exhibit intensity variations from sequence differences, prescan normalization, and acquisition-dependent gain changes. To handle intensity scaling between acquired $W$ and synthetic $\hat{W}$, we assume linear relationship $\hat{W}^{j}(x) = f(W^{j}) = a \cdot W^{j} + b$ where $a$ and $b$ are global scalar factors. Hence, rather than forcing absolute intensities to match, we maximize the correlation ratio~\cite{bib_b9} between $W^j$ and $\hat{W}^j$

\begin{equation}
    \eta^2(W^i \mid \hat{W}^i) = 1 - \frac{\text{Var}(W^i - \hat{W}^i)}{\text{Var}(W^i)},
\end{equation}
where $\text{Var}(\cdot)$ denotes the sample variance computed over brain-masked voxels. The guide loss as the complement of this ratio is
\begin{equation}
\begin{aligned}
    L_h(W, \hat{W}) =L_h
    &=\sum_{j}  \frac{\text{Var}(W^j - \hat{W}^j)}{\text{Var}(W^j)}.
    \label{ll}
\end{aligned}
\end{equation}

In summary, \eqref{eq:neg_log_likelihood} defines the objectives of SWIG qMRI SR, as: (1) Ensuring consistency with \( C_l \) by enforcing that images synthesized from  $\hat{Q}_h$\ using \eqref{phmodel}, after undergoing downsampling/blurring as described in \eqref{partialsampling}, are consistent with \( C_l \), (2) Maximizing the correlation ratio between the guide \( W^j \) and the synthesized image from $\hat{Q}_h$\ in \eqref{phmodel1} for all guides .

The two loss terms operate on different scales and represent different measurements: $L_l$ measures pixel-wise intensity consistency via MSE, while $L_h$ measures correlation ratio. Moreover, since $L_h$ does not derive from a full statistical description of the guide images, the scale of this term must be balanced with respect to the contrast image loss $L_l$. To balance their contributions during optimization, we introduce a weighting factor, $\alpha$, which controls the relative influence of structural guidance from the guide versus quantitative data consistency. Hence we rewrite \eqref{eq:neg_log_likelihood} as
\begin{equation}
L_{total} =  L_l + \alpha L_h.
\label{eq:weighted_loss}
\end{equation}

$L_{total}$ is computed exclusively from acquired LR contrast images $C_l$ and HR guide images $W$ --- neither term involves $Q_h$ at any point during training. The framework is therefore self-supervised in the sense that it requires no HR qMRI ground truth for optimization: $Q_h$ is never a supervision target, and the optimization is driven entirely by the physics-informed consistency between predicted and observed measurements.

\begin{figure*}[!ht]
    \centering
    \includegraphics[width=\textwidth]{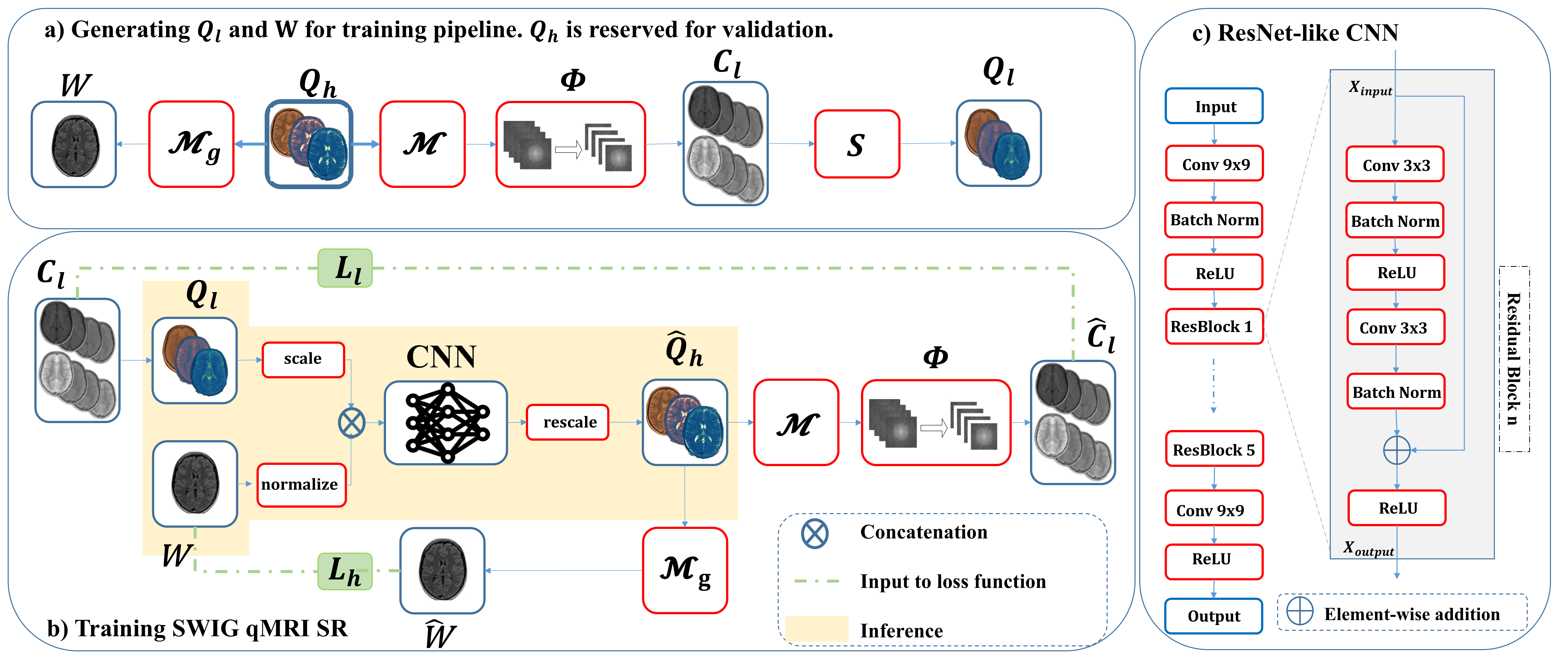}
    \caption{(a) Synthetic data generation for training the framework. (b) Training CNN and inference scheme for SWIG qMRI SR. (c) the ResNet-like CNN}
    \label{PHIREQ-training-scheme}
\end{figure*}

\subsection{Dataset}\label{subsec4}
We used two types of data: synthetic data for controlled experiments (training, validation, and ablation studies) and independently acquired in-vivo data to assess real-world performance and cross-sequence generalizability. All data acquisition protocols were approved by the institutional review board, and informed consent was obtained from all volunteers prior to scanning.

\subsubsection{Synthetic Data Generation}\label{subsubsec1}
For our experiments, we created a synthetic dataset from acquired qMRI scans. qMRI maps from 27 subjects were acquired using an 1.5T Optima MR450w and 1.5T SIGNA Artist (GE
Healthcare, Milwaukee, Wisconsin) with the AIR 24-channel head coil. The QRAPMASTER sequence~\cite{bib_b10} was used to obtain 2D parametric maps (\(Q_h\)) of proton density (PD), T1, and T2, with each slice reconstructed via MAGnetic resonance Image Compilation (MAGiC) software (SyntheticMR, Link\"oping, Sweden)~\cite{bib_b11}. The parameter maps have value ranges up to 160 arbitrary units (a.u.), 4300 ms, and 2000 ms for PD, T1, and T2, respectively.

To generate \( Q_l \), we synthesized 12 contrasts from \( Q_h \): 6 T1w and 6 T2w. The T1w contrasts used a saturation-recovery
\begin{equation}
    C_h^i = PD \cdot \left( 1 - e^{-TR^i/T1} \right), \quad i = 1, \dots, 6,
\end{equation}
with $TR^i \in \{360, 540, 810, 1215, 1822, 2733\}$ ms as repetition times. T2w contrasts used a spin-echo decay
\begin{equation}
    C_h^{i+6} = PD \cdot e^{-TE^i/T2}, \quad i = 1, \dots, 6,
\end{equation}
with $TE^i \in \{10, 20, 40, 80, 160, 320\}$ ms as echo times.

Then \( C_l^i \) was synthesized using \eqref{partialsampling} with complex Gaussian noise ($\sigma = 0.1$) added to the \( k\)-space, and a square binary mask \( B^i \), retaining one-quarter of central  \(k\)-space.
Maximum likelihood estimation was employed in \eqref{Signalfitting} to estimate \( Q_l \).

For \( W \), HR T1w synthesized using the inversion-recovery (IR) spoiled gradient-echo (SPGR) signal equation

\begin{align}
W^1 =\;&
    - M_{z,\mathrm{eq}} \cdot e^{-TD / T1}
    + PD \cdot \left( 1 - e^{-TD / T1} \right) \cdot e^{-TP / T1} \notag \\
    &+ PD \cdot \left( 1 - e^{-TP / T1} \right),
    \label{IR-GRE}
\end{align}
where the steady-state longitudinal magnetization \( M_{z,\mathrm{eq}} \) is given by

\begin{equation}
M_{z,\mathrm{eq}} = PD \cdot \frac{1 - e^{-TR / T1}}{1 - \cos(\alpha) \cdot e^{-TR / T1}} \cdot e^{-TE / T2},
\end{equation}
with \( TP \) as time from inversion to acquisition, \( TD \) as delay before inversion and \( \alpha \) as flip angle in degrees.

HR T2w synthesized using the spin-echo (SE) signal equation
\begin{equation}
W^2 = PD \cdot \frac{\sin(180^\circ - \alpha) \cdot \left(1 - e^{-(TR - TE)/T1} \right)}{1 - \cos(180^\circ - \alpha) \cdot e^{-(TR - TE)/T1}} \cdot e^{-TE/T2},
\label{PROPELLER}
\end{equation}

Acquisition parameters mimicking routine MRI were: T1w: \( TR/TE/TP/TD/\alpha = 6.6\,\mathrm{ms}/2.6\,\mathrm{ms}/450\,\mathrm{ms}/0 \mathrm{ms}/12^\circ\); T2w: \(
TR/TE/\alpha = 5211\mathrm{ms}/146\mathrm{ms}/160^\circ
\).

The dataset was split into 7 training, 3 validation, and 17 test subjects. All images were brain-masked using HD-BET~\cite{bib_b12}.
Note that although $Q_h$ is used to prepare training inputs, it is never used as a supervision target, consistent with the self-supervised formulation of $L_{total}$ in \eqref{eq:weighted_loss}. In these synthetic experiments, $Q_h$ is reserved exclusively as a held-out validation reference. Furthermore, because paired $Q_l$/$Q_h$ data are available in this controlled setting, they can also serve as a proxy phantom for tissue-specific quantitative evaluation of the proposed framework. For tissue-specific evaluation, white matter (WM), gray matter (GM), and cerebrospinal fluid (CSF) masks were obtained using FSL FAST~\cite{bib_b13}. Voxels were assigned to a tissue class only if the corresponding partial-volume estimate exceeded 0.9.

\subsubsection{Acquired Data for Validation}\label{acquired_data}

To evaluate for real-world performance and cross-sequence generalizability, we acquired data using a different qMRI protocol while maintaining identical acquisition parameters for $W$  used for guidance.

\textbf{qMRI acquisition and reconstruction:} Three healthy volunteers were scanned on a GE SIGNA Premier 3.0T scanner (AIR 48-channel head coil) using the Multi-Parametric Zero Echo Time (MuPa-ZTE) sequence~\cite{bib_b14,bib_b15}, a silent, phyllotaxis 3D acquisition for simultaneous PD, T1 and T2 mapping. Three qMRI quality levels were generated from this acquisition: (1) \textbf{Fast protocol (Acq1):} 1-minute acquisition simulated by retaining 20\% of acquired spokes, reconstructed with sparse fine gridding~\cite{bib_b16} and dictionary-based parameter fitting~\cite{bib_b17,bib_b18}, (2) \textbf{Standard protocol (Acq5):} 5-minute acquisition with sparse fine gridding and dictionary-based parameter fitting and (3) \textbf{ReconDL-enhanced protocol
(Acq5$^{\text{R}}$):} 5-minute acquisition reconstructed with
AIR\textsuperscript{TM} Recon DL~\cite{bib_b19,bib_b20} (supervised
convolutional neural network trained on HR references).
Acq5$^{\text{R}}$ is included as a vendor-provided supervised DL reconstruction representative of the type of learned reconstruction increasingly available in clinical scanners, making it a practically relevant reference point for benchmarking.

Each protocol serves as LR input to SWIG qMRI SR, while higher-quality protocols serve as reference. The pseudorandom phyllotaxis trajectory ensures uniform k-space sampling, making the fast protocol truncation representative of actual undersampled acquisitions~\cite{bib_b14,bib_b15,bib_b21}. This experimental design enables: (1)  performance evaluation of SWIG qMRI SR on independently acquired in-vivo data (2) whether guide images provide complementary information by comparing SR outputs from Acq1 and Acq5 inputs against the Acq5$^{\text{R}}$ reference, and (3) whether SR can further enhance maps already processed by state-of-the-art supervised DL reconstruction (Acq5$^{\text{R}}$ as input).

\textbf{Weighted images:} Standard clinical images (T1w IR-SPGR and T2w SE) were acquired with parameters matching the HR wMRI in the training dataset. A T2-FLAIR image was also acquired to evaluate synthesis fidelity for contrasts not used as guidance. All weighted images were registered and resampled to the qMRI coordinate system using rigid-body registration with elastix~\cite{bib_b22,bib_b23}.

\subsection{Implementation Details}\label{subsec5}

\subsubsection{Network Architecture}\label{subsubsec2}

The training objective is agnostic to the neural network (NN) architecture.
To evaluate the impact of the network design, two CNN models were
investigated under identical training conditions. As the primary model, we used a ResNet-based architecture~\cite{bib_b24}
with approximately 0.41M trainable parameters, consisting of an initial convolutional layer, five residual blocks, and a final output convolution (Fig.~\ref{PHIREQ-training-scheme}c). Each residual block contains two convolutional layers with batch normalization and ReLU activation connected through skip connections to facilitate gradient propagation. The final layer is a $9\times9$ convolution with linear activation to produce the refined HR parametric maps. As the secondary model, we used a U-Net architecture~\cite{bib_b25} with approximately 0.52M trainable parameters. The network consists of a two-level encoder--decoder structure with skip connections, residual blocks identical to those used in the ResNet model, a constant 64-channel width throughout the network, and $2\times2$ max-pooling and bilinear upsampling operations.

\subsubsection{Input Configuration and Preprocessing}\label{subsubsec3}
The network input consists of the LR qMRI, $Q_l$, (with one channel per parameter, e.g., $T_1$ and $T_2$ maps) concatenated channel-wise with the HR guide $W$ (Fig.~\ref{PHIREQ-training-scheme}b).

The qMRI maps have constant physiological ranges and must not be dynamically normalized to preserve quantitative accuracy. However, to provide the ResNet with appropriately scaled input values, parameter-specific scaling layer divide each channel of $Q_l$ by its mean value computed over brain-masked regions across the training dataset. The inverse scaling is applied to the output $\hat{Q}_h$ to restore correct quantitative values. The guide images $W$ may be arbitrarily scaled due to acquisition-dependent factors. Hence, each guide is z-score normalized to zero mean and unit variance, with statistics computed over brain-masked voxels. This normalization ensures stable training across parameters with different value scales.

\subsubsection{Training Configuration}\label{subsubsec4}

The model is implemented in PyTorch and trained to minimize the loss function $L_{total}$ in \eqref{eq:weighted_loss} using the ADAM optimizer with a learning rate of $10^{-5}$ on an A40 GPU (NVIDIA, Santa Clara, CA).
The training scheme is shown in Fig.~\ref{PHIREQ-training-scheme}b.

\subsection{Evaluation Metrics}\label{sec:eval_metrics}
We evaluate SR outputs using direct qMRI assessment and indirect evaluation via synthetic weighted image quality.

\subsubsection{Direct qMRI Quality Assessment}\label{subsubsec5}
For experiments with available ground-truth HR qMRI ($Q_h$), we compute:
(1) Structural Similarity Index (SSIM)~\cite{bib_b26}: which approximates perceptual similarity (range 0--1, higher is better),
(2) High-Frequency Error Norm (HFEN)~\cite{bib_b27}: which quantifies edge preservation errors (lower is better) and (3) tissue-specific Mean Absolute Error (MAE): which measures voxel-wise quantitative accuracy in physical units (ms for T1 and T2, a.u.\ for PD; lower is better). For map $j$, MAE is defined as
\begin{equation}
\mathrm{MAE}_j = \frac{1}{|\Omega|} \sum_{x \in \Omega} \left| \hat{Q}_{h,j}(x) - Q_{h,j}(x) \right|,
\label{eq:mae}
\end{equation}
where $\Omega$ denotes the brain-masked evaluation region.
Metrics are computed between $\hat{Q}_h$ and $Q_h$, while the metric computed between $Q_l$ and $Q_h$ is provided as baseline.

\subsubsection{Indirect Assessment via Synthetic Weighted Images}\label{subsubsec6}
Since wMRIs are the primary diagnostic modality in clinical practice,
we assess SR performance quality by evaluating synthesized wMRI
realism. For each image in $W$, we: (1) synthesize corresponding
images from $Q_l$ and $\hat{Q}_h$ using forward models with
acquisition-matched parameters, (2) compute SSIM, HFEN, and
Peak Signal-to-Noise Ratio (PSNR), which measures overall
reconstruction fidelity in dB (higher is better), between synthetic and acquired images, and (3) compare synthesis quality from $\hat{Q}_h$ versus from $Q_l$.

\subsection{Experimental Validation}\label{subsec6}

\subsubsection{Ablation Study: Individual Component Contributions}\label{Ablation Study: Individual Component Contributions}
We systematically evaluate the contribution of each framework component through controlled ablation experiments. Specifically, we address two questions: (1) Can HR wMRIs (T1w, T2w) effectively guide qMRI SR without HR qMRI training data? (2) Is LR qMRI necessary at inference, or can SR be performed using only HR wMRI?

To answer these questions, we train eight model variants with different input ,supervision and NN configurations (Table~\ref{tab:model-configs}). We denote these models as $M_{guide}^{qMRI}$ where the subscript indicates the guide (1=T1w, 2=T2w, 12=both) and superscript positions encode: (1) whether $Q_l$ is model input (+/-), (2) whether $C_l$ data consistency is used in loss (+/-) and if the secondary NN backbone, U-Net, is used (U).

Model $M_{12}^{++}$ represents the full SWIG qMRI SR framework with all components active. Models $M_{1}^{++}$ and $M_{2}^{++}$ isolate the contribution of individual guide. Model $M_{12}^{-+}$ tests whether HR wMRIs alone can provide sufficient information when data consistency is maintained during training but $Q_l$ is unavailable at inference. Model $M_{12}^{--}$ represents qMRI mapping from wMRI without any quantitative constraints. Together, $M_{12}^{-+}$ and $M_{12}^{--}$ probe whether two weighted images (one T1w and one T2w)---a constraint representative of direct wMRI-to-qMRI mapping approaches and most clinical protocols---are sufficient for accurate quantitative parameter recovery in the absence of LR qMRI as an anchor.

To contextualize our framework against direct wMRI-to-qMRI approaches previously proposed by Qiu et al.~\cite{bib_b5}, we implement an adapted variant of their formulation, denoted $M_{12}^{-+,Qiu}$, under our experimental conditions. This variant retains the same weighted-image signal models, network architecture, and training data used throughout this work, and additionally includes a synthetic FLAIR guide, generated
using the inversion-recovery spin-echo signal equation, consistent with the original method's input configuration; similar to the original method, this synthetic FLAIR is used only as an additional guide input and not as a supervision target. We also replace the guide (HR) loss term with that of Qiu et al. -- a Sobel-edge-based loss -- in place of our correlation-ratio guide loss, while retaining our own physics-informed LR consistency term $L_l$, rather than adopting the direct MSE supervision of $Q_l$ used in the original method. This isolates the effect of the guide-loss formulation while holding the LR consistency mechanism, network architecture, input configuration, and training data constant. Aside from the additional FLAIR guide and the modified guide
loss, $M_{12}^{-+,Qiu}$ otherwise shares the same $Q_l$-input configuration as $M_{12}^{-+}$ (i.e., no LR qMRI input at inference).

We also define model $M^{s}$ which serves as a fully supervised upper bound: the network architecture and inputs---$Q_l$ and  $W_h$ guides---are identical to $M_{12}^{++}$, but the physics-informed losses $L_l$ and $L_h$ are replaced by direct MSE supervision against $Q_h$. Since $Q_h$ is available in the synthetic experiment, this model quantifies the performance ceiling achievable with HR qMRI supervision, and any gap between $M^{s}$ and $M_{12}^{++}$ reflects the cost of the self-supervised formulation. Note that $M^{s}$ requires HR qMRI as a training target and therefore cannot be deployed in the clinical scenario that motivates SWIG qMRI SR.

\begin{table}[!h]
\centering
\caption{Model configurations for ablation experiments. Notation:
$M_{guide}^{qMRI}$ where subscript=guide (1=T1w, 2=T2w, 12=both);
superscript first position=$Q_l$ as input (+=yes, -=no); second
position=$L_l$ data consistency (+=yes, -=no). $M_{12}^{++,U}$ uses a U-Net backbone under identical conditions to $M_{12}^{++}$.
$M^{s}$ denotes the fully supervised upper bound. $M_{12}^{-+,Qiu}$ denotes the adapted Qiu et al.~\cite{bib_b5} variant.}
\label{tab:model-configs}
\renewcommand{\arraystretch}{2}
\vspace{2mm}
\begin{tabularx}{\columnwidth}{%
  >{\hsize=0.8\hsize}Y
  >{\hsize=1.0\hsize}Y
  >{\hsize=1.2\hsize}Y
  >{\hsize=1.0\hsize}Y
  >{\hsize=1.0\hsize}Y
}
\toprule
Model & qMRI Input & Loss & T1w Guide & T2w Guide \\
\midrule
\( M_{12}^{++} \)             & Yes & $L_l + \alpha L_h$ & Yes & Yes \\
\( M_{2}^{++} \)              & Yes & $L_l + \alpha L_h$ & No  & Yes \\
\( M_{1}^{++} \)              & Yes & $L_l + \alpha L_h$ & Yes & No  \\
\( M_{12}^{-+} \)             & No  & $L_l + \alpha L_h$ & Yes & Yes \\
\( M_{12}^{--} \)             & No  & $\alpha L_h$               & Yes & Yes \\
\( M_{12}^{++,\text{U}} \) & Yes & $L_l + \alpha L_h$ & Yes & Yes \\
\( M_{12}^{-+,Qiu} \) & No & $L_l + \alpha L'_h$ & Yes & Yes \textsuperscript{*} \\
\( M^{s} \)           & Yes & MSE$(Q_h)$          & Yes & Yes \\
\bottomrule
\end{tabularx}

\par
\vspace{1mm}
{\raggedright
\textsuperscript{*}$M_{12}^{-+,Qiu}$ additionally uses a synthetic FLAIR guide as input, matching the original input configuration of Qiu et al. $L'_h$ denotes the Sobel-edge guide loss used in place of $L_h$ in this variant.
\par}

\end{table}
\textbf{Sensitivity Analysis of Loss Weighting Parameter $\alpha$:} To assess the sensitivity of the framework to the loss weighting parameter $\alpha$, we trained eight independent instances of $M_{12}^{++}$ with $\alpha \in \{0.01, 0.1, 1, 10, 100, 1000, 10000, 100000\}$, holding all other hyperparameters fixed. Evaluation was performed on the validation subjects using SSIM and HFEN computed over the entire brain-masked region, as well as the tissue-specific MAE (\eqref{eq:mae}) for each parametric map (PD, T1, and T2). The $\alpha$ value minimizing the mean MAE across the three maps was selected and used for all subsequent experiments reported in this work.

\textbf{Model Selection:} We rank ablation models trained with chosen $\alpha$ separately for each parametric map (PD, T1, T2) based on mean SSIM and HFEN values across 17 test subjects, assigning lower ranks to better performance
(higher SSIM, lower HFEN). Models $M_{12}^{++,U}$, $M^{s}$, and $M_{12}^{-+,Qiu}$ are excluded from ranking as they serve distinct purposes: $M_{12}^{++,U}$ evaluates NN architectural sensitivity, $M^{s}$
establishes a supervised upper bound that cannot be deployed in the clinical setting that motivates this work, and $M_{12}^{-+,Qiu}$ serves as a baseline comparison against an adapted prior work. The final model score is the sum of ranks across both metrics and maps. The lowest-scoring model is selected for validation on separately acquired in-vivo data.

\subsubsection{Evaluation on In-vivo Data}\label{subsubsec7}
In this section we apply the best-performing model to three protocols described in Section~\ref{acquired_data}. The same model trained on the synthetic data with $\alpha=1000$ was applied to in-vivo data without retraining or fine-tuning.

\textbf{qMRI Qualitative evaluation:} We compare ($\hat{Q}_h$) qualitatively against higher quality qMRI, focusing on structural detail recovery and quantitative value preservation. Additionally, by treating qMRI from ReconDL-enhanced protocol as  $Q_l$  we test whether our framework can provide further improvement beyond what a state-of-the-art supervised DL reconstruction alone achieves.

\textbf{Evaluation via wMRI Synthesis:} Since qMRI are valuable for their ability to synthesize contrasts via forward models, we validate SR quality by assessing synthesis fidelity. If $\hat{Q}_h$ accurately represents underlying tissue properties, it should synthesize wMRI that closely match acquired clinical scans. We evaluate synthesis quality in two complementary ways: (1) For each guide $W^i$, we synthesize corresponding images from both $Q_l$ and $\hat{Q}_h$ using the corresponding forward models (\eqref{IR-GRE}, \eqref{PROPELLER}) with acquisition-matched parameters. (2) We synthesize T2-FLAIR from $\hat{Q}_h$  using inversion-recovery signal model

\begin{align}
\text{T2-FLAIR} =\;
     \left(1 - 2e^{-TI / T1} + e^{-TR / T1} \right)\cdot e^{-TE / T2} ,
    \label{T2-FLAIR-signal-equation}
\end{align}
where $TI$ is the inversion time and acqusition parameters matched to the acquired T2-FLAIR sequence, and compare against an independently acquired T2-FLAIR scan that was never used as guidance.

Synthesis quality is assessed both quantitatively (SSIM, HFEN and PSNR against acquired wMRI) and qualitatively (visual comparison of fluid suppression quality, anatomical detail, and artifacts). Superior synthesis fidelity from $\hat{Q}_h$ compared to $Q_l$ would indicate that SR has enhanced the quantitative information content beyond the original LR maps.

The second test evaluates whether $\hat{Q}_h$ capture sufficient quantitative tissue information to predict contrasts outside the training distribution. Successful T2-FLAIR synthesis would demonstrate that SWIG qMRI SR transfers structural information from guide into quantitatively meaningful and correct parametric maps rather than merely copying guide features. Synthesized T2-FLAIR from $Q_l$ establishes baseline for qMRI SR.

\begin{figure*}[!h]
    \centering
    \includegraphics[width=\textwidth]{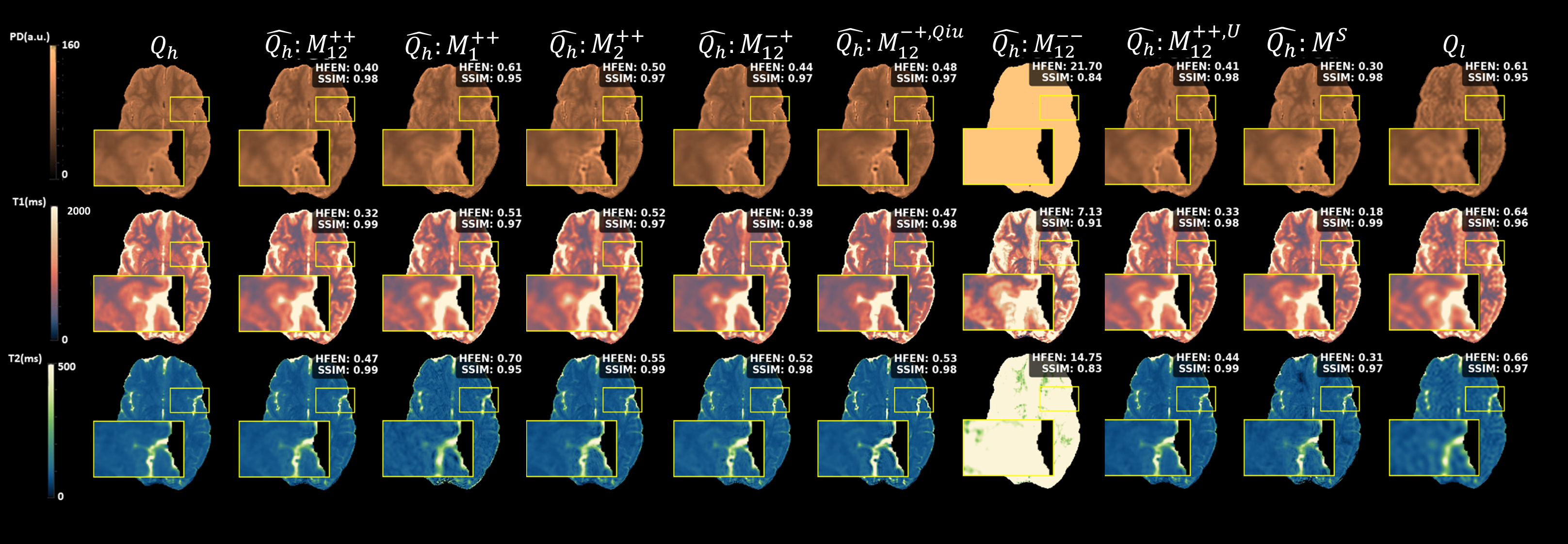}
    \caption{Super-resolved parametric maps from ablation study experiments on a representative slice. Rows: PD (top), T1 (middle), T2 (bottom). Columns: ground-truth $Q_h$ (left), model predictions (center), LR input $Q_l$ (right). Slice-specific SSIM and HFEN relative to $Q_h$ shown above each map}
    \label{fig:synthetic_data_Slice}
\end{figure*}

\section{Results}\label{sec4}

\begin{figure*}[!h]
    \centering
    \includegraphics[width=\textwidth]{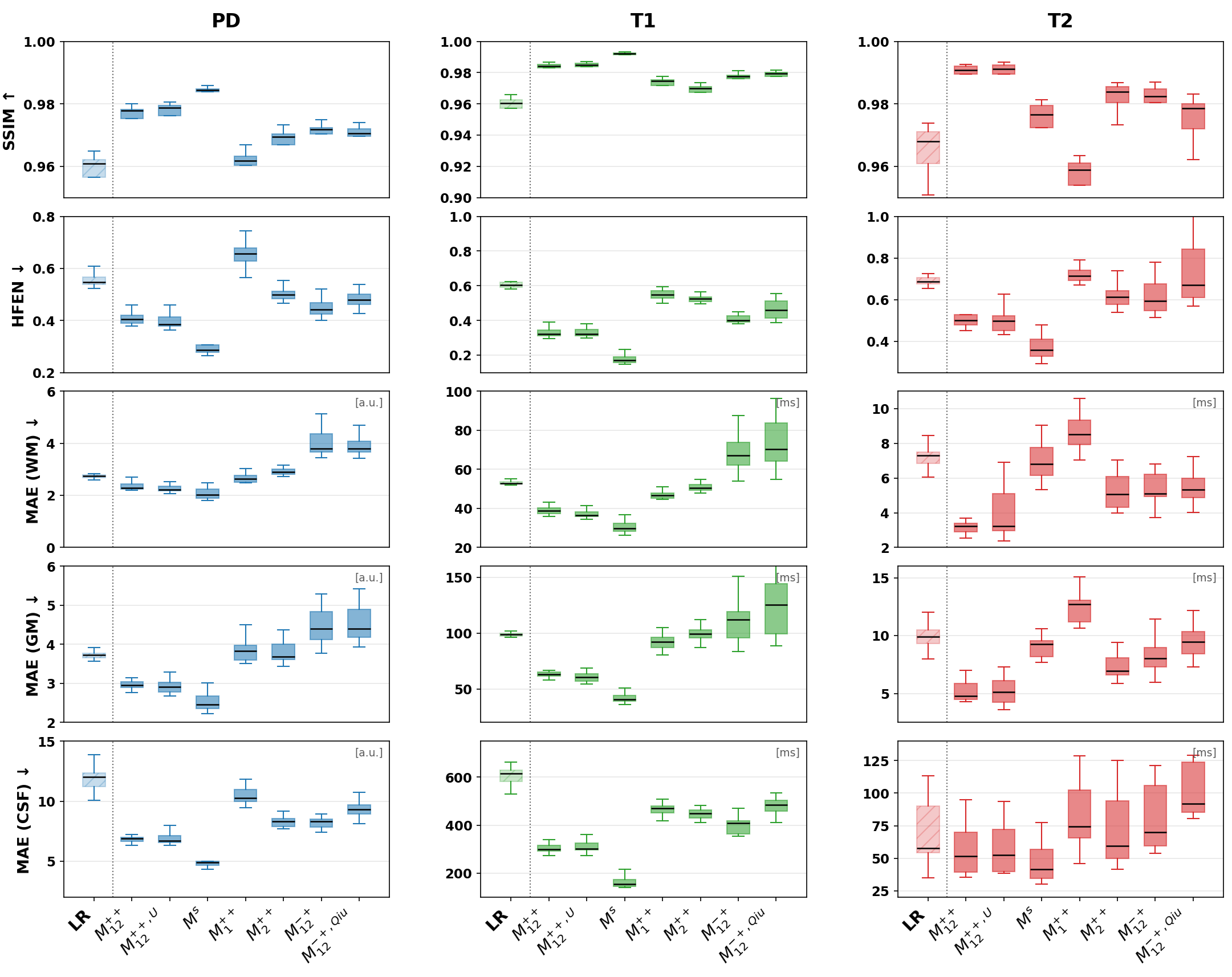}
  \caption{Quantitative SR evaluation across 17 test subjects for different SWIG qMRI SR model variants in the ablation. Top rows: SSIM (higher is better) and HFEN (lower is better) for PD (blue), T1 (green), and T2 (red) maps. Bottom rows: mean absolute error (MAE, lower is better) evaluated within white matter (WM), gray matter (GM), and CSF regions. Boxplots summarize per-subject metrics averaged over brain-masked slices. LR denotes the low-resolution baseline $Q_l$ versus $Q_h$}
    \label{fig:boxplots}
\end{figure*}

\begin{figure}[!h]
    \centering
    \includegraphics[width=\columnwidth]{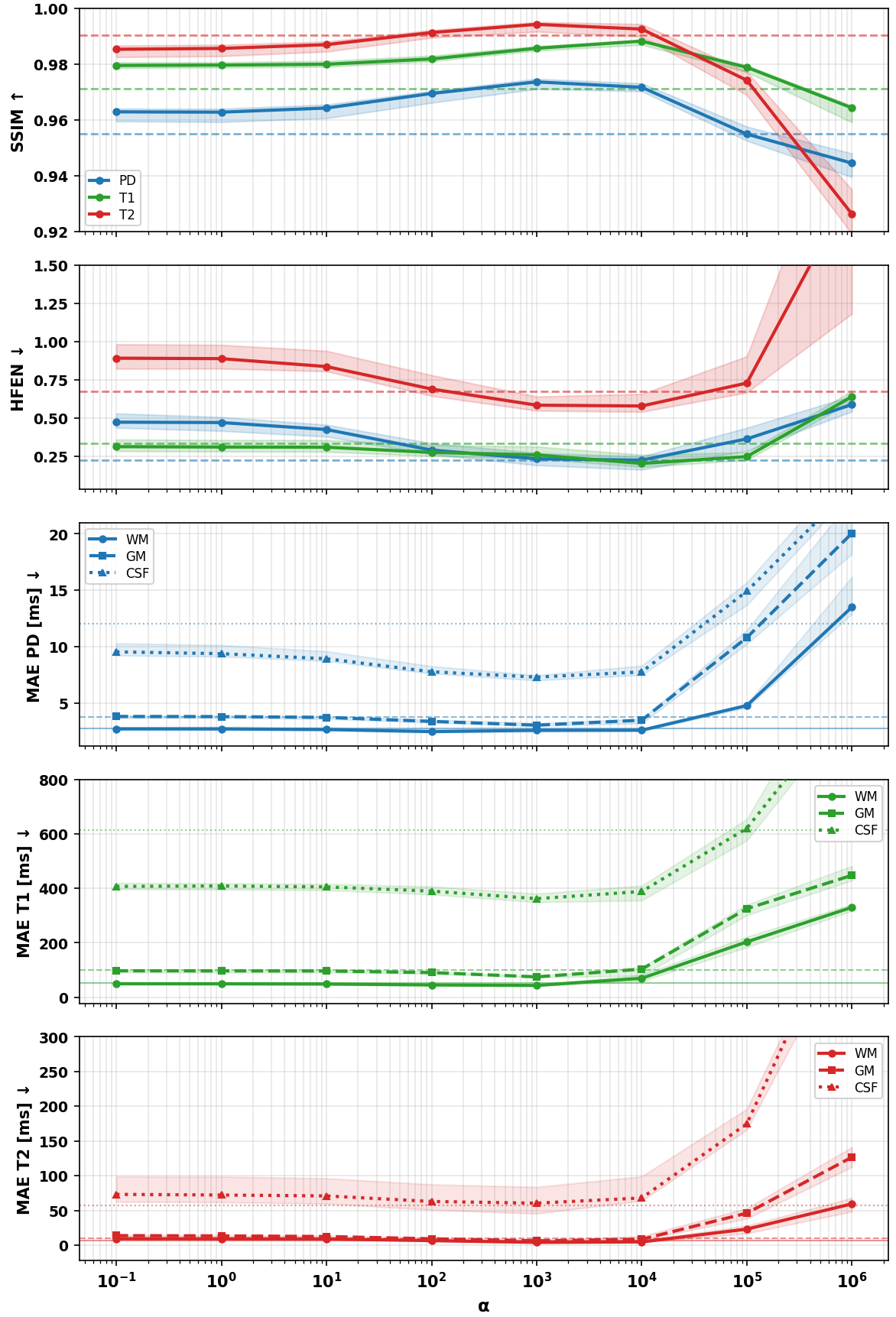}
\caption{Sensitivity of SR performance to $\alpha$ across $N=17$
healthy volunteers. Solid lines: median; shaded bands:
interquartile range; dashed lines: LR baseline. Top two rows: SSIM
and HFEN for PD (blue), T1 (green), and T2 (red). Bottom three
rows: tissue-specific MAE with WM (solid), GM (dashed), and CSF
(dotted) lines}
    \label{fig:alpha_sensitivity}
\end{figure}

\subsection{Ablation Study: Individual Component Contributions}\label{subsec7}
\textbf{Qualitative Performance on Single Slice:} Figure~\ref{fig:synthetic_data_Slice} presents qMRI SR results from eight SR model variants. Model $M_{1}^{++}$ enhanced T1 maps while minimally affecting T2 maps. Conversely, $M_{2}^{++}$ improved T2 maps markedly with some T1 impact.
The dual-guided $M_{12}^{++}$ achieved balanced enhancement across all
parameters. Model $M_{12}^{-+}$ (no qMRI input at inference) produced qMRI maps with quantitative errors, particularly in gray matter T1 and PD estimation, while SSIM and HFEN improved for all maps. Model $M_{12}^{-+,Qiu}$ showed similar behavior to $M_{12}^{-+}$. Model $M_{12}^{--}$ failed completely and was excluded from further analysis.
The variant $M_{12}^{++,U}$ produced visually comparable results to $M_{12}^{++}$ and comparable metrics for this slice, while $M^{s}$ achieved the best metrics across all maps.

\textbf{Quantitative Performance Across Test Subjects:}
Figure~\ref{fig:boxplots} summarizes SSIM, HFEN, and tissue-specific MAE across the 17 test subjects for seven model variants. Compared with the LR baseline, the full dual-guided model $M_{12}^{++}$ improved SSIM and HFEN across all three parameter
maps. The largest SSIM gain was observed for PD, increasing from ${\sim}0.960$ to ${\sim}0.977$, while T1 HFEN showed the largest
absolute reduction, decreasing from ${\sim}0.60$ to ${\sim}0.33$. Tissue-specific MAE was also reduced across PD, T1, and T2 in WM,
GM, and CSF. Notably, T1 GM MAE decreased from ${\sim}100$\,ms to ${\sim}63$\,ms, while PD GM MAE decreased from ${\sim}3.8$ to
${\sim}3.0$ in arbitrary units.

The single-guide models exhibited parameter-specific behavior. $M_{1}^{++}$, using only the T1w guide, achieved lower T1 MAE than $M_{2}^{++}$ across all tissue types, whereas $M_{2}^{++}$, using only the T2w guide, yielded lower T2 MAE. The dual-guided
$M_{12}^{++}$ model generally provided a more balanced performance across all three parameter maps.

Removing the LR qMRI input degraded quantitative accuracy. Although $M_{12}^{-+}$ retained SSIM values above the LR baseline,
it exhibited higher HFEN and tissue-specific MAE than $M_{12}^{++}$. The degradation was particularly pronounced for T1,
with GM MAE increasing from ${\sim}63$ to ${\sim}110$\,ms and WM MAE increasing from ${\sim}39$ to ${\sim}67$\,ms. This indicates that the weighted-image guides alone provide useful structural information, but the LR qMRI input remains important for preserving
quantitative fidelity. The adapted Qiu-style model, $M_{12}^{-+,Qiu}$, also improved upon
the LR baseline for most metrics, but generally underperformed $M_{12}^{-+}$. The two models achieved comparable PD SSIM, whereas
$M_{12}^{-+,Qiu}$ showed lower T2 SSIM and higher HFEN across all three parameter maps. The Qiu-style formulation also produced
higher MAE in several tissue--parameter combinations, most notably for T1 in GM and CSF and for T2 in GM and CSF.

The U-Net variant $M_{12}^{++,\mathrm{U}}$ achieved SSIM and HFEN values comparable to those of $M_{12}^{++}$, with similar or
slightly lower tissue-specific MAE in several regions, including T1 WM and GM.

The supervised upper bound $M^{s}$ achieved the lowest MAE for PD and T1 across all tissue types all three parameter maps.

\textbf{Model Selection:}
Table~\ref{tab:improvement} gives the mean percentage improvements in metric values across all subjects. Ranking models by performance on each metric-map combination (lower rank = better), $M_{12}^{++}$ achieved the lowest total rank score (6) and hence, was selected for further validation.

\textbf{Sensitivity Analysis of Loss Weighting Parameter
$\alpha$:} Figure~\ref{fig:alpha_sensitivity} shows SSIM, HFEN, and
tissue-specific MAE as a function of $\alpha$. For low values
($\alpha \leq 100$), performance remained relatively stable across
all metrics. SSIM and HFEN improved between $\alpha = 10^{2}$ and
$10^{4}$, with the best overall trade-off observed at
$\alpha = 10^{3}$. Beyond $\alpha = 10^{4}$, all metrics degraded
sharply. Based on this analysis, $\alpha = 1000$ was
selected for all other experiments in this study.

\begin{table*}[!t]
\centering
\caption{Ablation study performance: percentage improvement over
$Q_l$ baseline for each model and parametric map (averaged across 17
test subjects). Total rank was computed by ranking the ablation models
for each SSIM and HFEN entry separately and summing ranks across all six metric--map combinations (lower total rank indicates better overall performance).}
\label{tab:improvement}

\setlength{\tabcolsep}{3pt}
\renewcommand{\arraystretch}{1.1}

\scriptsize

\begin{tabularx}{\textwidth}{ll*{7}{Y}}
\toprule
\textbf{Metric} &
\textbf{Map} &
$\mathbf{M_{1}^{++}}$ &
$\mathbf{M_{2}^{++}}$ &
$\mathbf{M_{12}^{++}}$ &
$\mathbf{M_{12}^{-+}}$ &
$\mathbf{M_{12}^{++,\mathrm{U}}}$ &
$\mathbf{M^{s}}$ &
$\mathbf{M_{12}^{-+,\mathrm{Qiu}}}$ \\
& & (\%) & (\%) & (\%) & (\%) & (\%) & (\%) & (\%) \\
\midrule

\multirow{3}{*}{SSIM}
 & PD & +0.30 & +0.96 & +2.00 & +1.06 & +2.10 & +2.79 & +0.95 \\
 & T1 & +1.60 & +1.00 & +2.81 & +1.88 & +2.92 & +3.84 & +2.20 \\
 & T2 & -1.32 & +1.91 & +2.89 & +1.80 & +2.91 & +1.14 & +1.41 \\
\midrule

\multirow{3}{*}{HFEN}
 & PD & -16.18 & +9.37 & +25.84 & +19.11 & +28.32 & +45.73 & +12.72 \\
 & T1 & +9.41  & +13.28 & +45.39 & +32.66 & +45.12 & +69.82 & +23.31 \\
 & T2 & -3.11  & +10.17 & +24.10 & +11.96 & +26.11 & +47.22 & -2.26 \\
\midrule

\multicolumn{2}{l}{\textbf{Total rank}}
 & 23 & 18 & \textbf{6} & 13 & -- & -- & -- \\
\bottomrule
\end{tabularx}
\end{table*}

\begin{figure*}[!t]
   \centering
   \includegraphics[width=\textwidth]{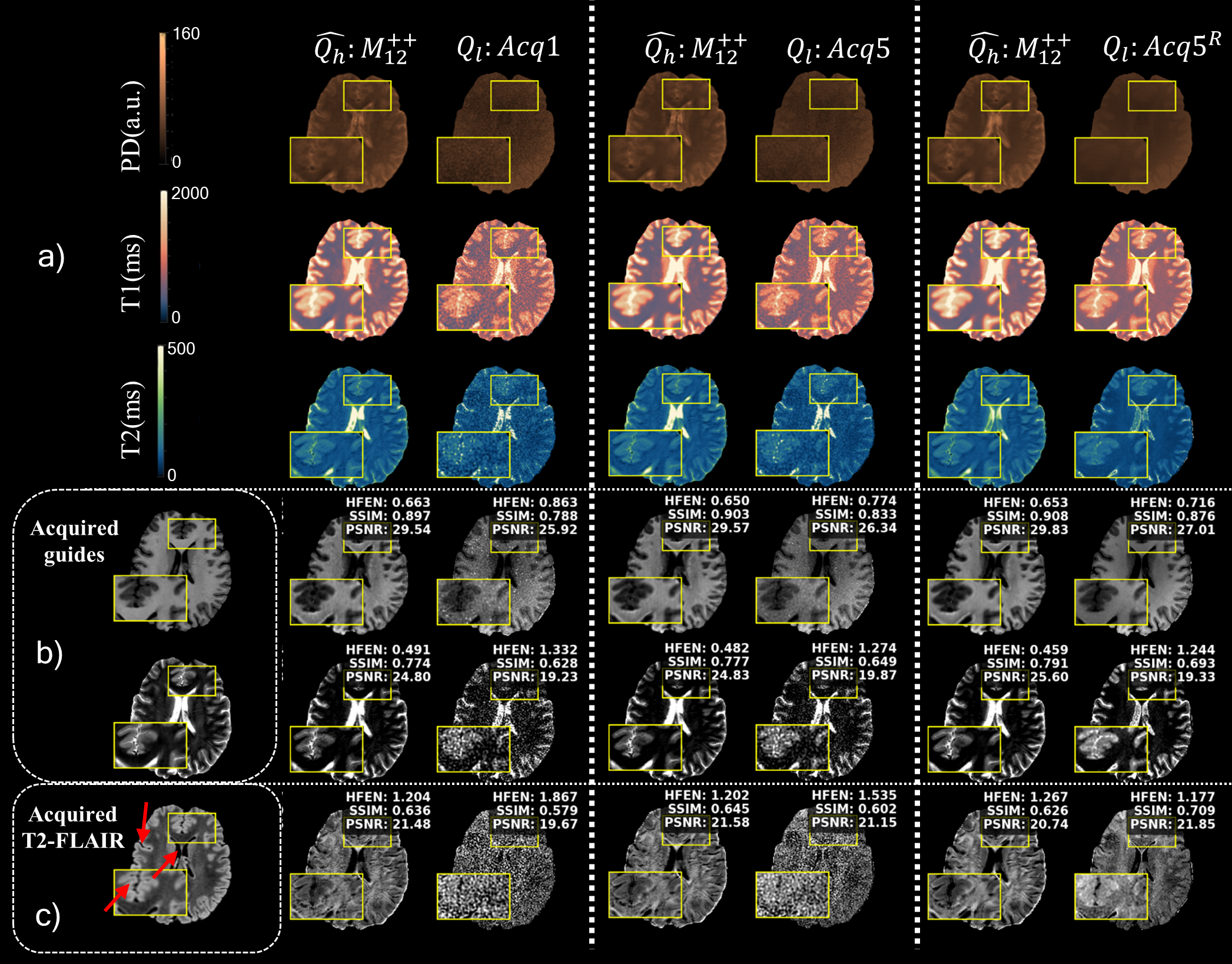}
    \caption{(a) Model $M_{12}^{++}$ applied to MuPa-ZTE volunteer 1. Rows: PD, T1, T2 maps. Columns per protocol (Acq1, Acq5, Acq5$^{\text{R}}$): input $Q_l$, SR output $\hat{Q}_h$. (b) Acquired guides (left) and synthesized images from $Q_l$ and $\hat{Q}_h$ using forward models with acquisition-matched parameters of acquired guides. (c) T2-FLAIR synthesized from $Q_l$ and $\hat{Q}_h$ (never used as guide) versus acquired reference. Metrics in b and c computed for the shown slice}
   \label{fig:slice_acquired_data}
\end{figure*}

\begin{table*}[!h]
\centering
\scriptsize
\renewcommand{\arraystretch}{1.12}
\setlength{\tabcolsep}{2pt}
\caption{Subject-wise quantitative evaluation of wMRI synthesized from
${Q}_l$ and $\hat{Q}_h$ with respect to acquired clinical references.
For each subject, metrics are reported as mean$\pm$std across axial slices
40--140; PSNR in dB. FLAIR denotes T2-FLAIR. SSIM and PSNR: higher is
better; HFEN: lower is better. Bold indicates the numerically better mean
value according to the metric direction.}
\label{tab:metrics_inference_subjectwise}
\begin{tabular*}{\textwidth}{@{\extracolsep{\fill}}cclcccccc@{}}
\toprule
& & & \multicolumn{2}{c}{SSIM ($\uparrow$)}
& \multicolumn{2}{c}{PSNR ($\uparrow$)}
& \multicolumn{2}{c}{HFEN ($\downarrow$)} \\
\cmidrule(lr){4-5}\cmidrule(lr){6-7}\cmidrule(lr){8-9}
& & Contrast
& $Q_l$ & $\hat{Q}_h$ & $Q_l$ & $\hat{Q}_h$ & $Q_l$ & $\hat{Q}_h$ \\
\midrule

\multirow{9}{*}{\rotatebox[origin=c]{90}{Subject 1}}
& \multirow{3}{*}{\rotatebox[origin=c]{90}{Acq1}}
& T1w   & 0.82$\pm$0.05 & \textbf{0.91$\pm$0.05}
        & 25.1$\pm$0.6 & \textbf{27.3$\pm$0.5}
        & 0.75$\pm$0.10 & \textbf{0.35$\pm$0.05} \\
& & T2w & 0.77$\pm$0.08 & \textbf{0.87$\pm$0.05}
        & 19.3$\pm$0.9 & \textbf{24.5$\pm$0.7}
        & 0.89$\pm$0.11 & \textbf{0.29$\pm$0.05} \\
& & FLAIR & 0.69$\pm$0.09 & \textbf{0.75$\pm$0.07}
        & 19.3$\pm$1.0 & \textbf{22.0$\pm$0.8}
        & 1.20$\pm$0.10 & \textbf{1.09$\pm$0.14} \\
\cmidrule(lr){2-9}

& \multirow{3}{*}{\rotatebox[origin=c]{90}{Acq5}}
& T1w   & 0.83$\pm$0.05 & \textbf{0.92$\pm$0.05}
        & 25.4$\pm$0.6 & \textbf{27.2$\pm$0.4}
        & 0.73$\pm$0.09 & \textbf{0.34$\pm$0.05} \\
& & T2w & 0.74$\pm$0.10 & \textbf{0.87$\pm$0.06}
        & 19.9$\pm$0.9 & \textbf{24.6$\pm$0.7}
        & 1.09$\pm$0.09 & \textbf{0.28$\pm$0.05} \\
& & FLAIR & 0.67$\pm$0.10 & \textbf{0.74$\pm$0.08}
        & 20.8$\pm$0.9 & \textbf{22.1$\pm$0.8}
        & 1.38$\pm$0.08 & \textbf{1.08$\pm$0.13} \\
\cmidrule(lr){2-9}

& \multirow{3}{*}{\rotatebox[origin=c]{90}{Acq5$^{\mathrm{R}}$}}
& T1w   & 0.85$\pm$0.04 & \textbf{0.93$\pm$0.03}
        & 25.7$\pm$0.5 & \textbf{27.2$\pm$0.4}
        & 0.69$\pm$0.09 & \textbf{0.48$\pm$0.07} \\
& & T2w & 0.61$\pm$0.13 & \textbf{0.84$\pm$0.07}
        & 20.2$\pm$0.8 & \textbf{25.0$\pm$0.6}
        & 1.42$\pm$0.15 & \textbf{0.39$\pm$0.07} \\
& & FLAIR & \textbf{0.68$\pm$0.10} & 0.60$\pm$0.08
        & \textbf{21.0$\pm$0.9} & 19.1$\pm$0.8
        & 1.58$\pm$0.17 & \textbf{1.41$\pm$0.10} \\

\midrule

\multirow{9}{*}{\rotatebox[origin=c]{90}{Subject 2}}
& \multirow{3}{*}{\rotatebox[origin=c]{90}{Acq1}}
& T1w   & 0.84$\pm$0.04 & \textbf{0.95$\pm$0.03}
        & 25.5$\pm$0.6 & \textbf{27.6$\pm$0.5}
        & 0.71$\pm$0.09 & \textbf{0.33$\pm$0.04} \\
& & T2w & 0.75$\pm$0.09 & \textbf{0.86$\pm$0.06}
        & 19.6$\pm$0.9 & \textbf{24.3$\pm$0.8}
        & 0.94$\pm$0.12 & \textbf{0.31$\pm$0.06} \\
& & FLAIR & 0.71$\pm$0.08 & \textbf{0.77$\pm$0.06}
        & 19.7$\pm$1.0 & \textbf{22.3$\pm$0.8}
        & 1.16$\pm$0.11 & \textbf{1.04$\pm$0.12} \\
\cmidrule(lr){2-9}

& \multirow{3}{*}{\rotatebox[origin=c]{90}{Acq5}}
& T1w   & 0.84$\pm$0.04 & \textbf{0.94$\pm$0.02}
        & 25.6$\pm$0.6 & \textbf{27.5$\pm$0.4}
        & 0.70$\pm$0.08 & \textbf{0.32$\pm$0.04} \\
& & T2w & 0.72$\pm$0.11 & \textbf{0.86$\pm$0.06}
        & 20.2$\pm$0.8 & \textbf{24.4$\pm$0.7}
        & 1.14$\pm$0.10 & \textbf{0.30$\pm$0.05} \\
& & FLAIR & 0.69$\pm$0.09 & \textbf{0.76$\pm$0.07}
        & 21.1$\pm$0.9 & \textbf{22.4$\pm$0.8}
        & 1.32$\pm$0.09 & \textbf{1.03$\pm$0.12} \\
\cmidrule(lr){2-9}

& \multirow{3}{*}{\rotatebox[origin=c]{90}{Acq5$^{\mathrm{R}}$}}
& T1w   & 0.86$\pm$0.04 & \textbf{0.94$\pm$0.03}
        & 25.9$\pm$0.5 & \textbf{27.5$\pm$0.4}
        & 0.66$\pm$0.08 & \textbf{0.45$\pm$0.06} \\
& & T2w & 0.63$\pm$0.12 & \textbf{0.85$\pm$0.07}
        & 20.4$\pm$0.8 & \textbf{25.2$\pm$0.6}
        & 1.37$\pm$0.14 & \textbf{0.37$\pm$0.06} \\
& & FLAIR & \textbf{0.70$\pm$0.09} & 0.62$\pm$0.08
        & \textbf{21.3$\pm$0.9} & 19.4$\pm$0.8
        & 1.52$\pm$0.15 & \textbf{1.36$\pm$0.11} \\

\midrule

\multirow{9}{*}{\rotatebox[origin=c]{90}{Subject 3}}
& \multirow{3}{*}{\rotatebox[origin=c]{90}{Acq1}}
& T1w   & 0.81$\pm$0.06 & \textbf{0.92$\pm$0.04}
        & 24.8$\pm$0.7 & \textbf{27.0$\pm$0.5}
        & 0.79$\pm$0.11 & \textbf{0.37$\pm$0.06} \\
& & T2w & 0.78$\pm$0.08 & \textbf{0.88$\pm$0.05}
        & 19.0$\pm$1.0 & \textbf{24.8$\pm$0.7}
        & 0.86$\pm$0.10 & \textbf{0.27$\pm$0.05} \\
& & FLAIR & 0.68$\pm$0.10 & \textbf{0.74$\pm$0.08}
        & 19.1$\pm$1.1 & \textbf{21.7$\pm$0.9}
        & 1.24$\pm$0.12 & \textbf{1.13$\pm$0.15} \\
\cmidrule(lr){2-9}

& \multirow{3}{*}{\rotatebox[origin=c]{90}{Acq5}}
& T1w   & 0.82$\pm$0.06 & \textbf{0.92$\pm$0.04}
        & 25.1$\pm$0.6 & \textbf{26.9$\pm$0.5}
        & 0.77$\pm$0.10 & \textbf{0.36$\pm$0.06} \\
& & T2w & 0.76$\pm$0.09 & \textbf{0.88$\pm$0.05}
        & 19.7$\pm$0.9 & \textbf{24.9$\pm$0.7}
        & 1.04$\pm$0.08 & \textbf{0.26$\pm$0.04} \\
& & FLAIR & 0.66$\pm$0.11 & \textbf{0.73$\pm$0.09}
        & 20.6$\pm$1.0 & \textbf{21.8$\pm$0.8}
        & 1.43$\pm$0.09 & \textbf{1.12$\pm$0.14} \\
\cmidrule(lr){2-9}

& \multirow{3}{*}{\rotatebox[origin=c]{90}{Acq5$^{\mathrm{R}}$}}
& T1w   & 0.84$\pm$0.05 & \textbf{0.92$\pm$0.04}
        & 25.5$\pm$0.6 & \textbf{26.9$\pm$0.4}
        & 0.72$\pm$0.10 & \textbf{0.51$\pm$0.08} \\
& & T2w & 0.59$\pm$0.14 & \textbf{0.83$\pm$0.08}
        & 19.9$\pm$0.9 & \textbf{24.7$\pm$0.7}
        & 1.47$\pm$0.16 & \textbf{0.42$\pm$0.08} \\
& & FLAIR & \textbf{0.67$\pm$0.11} & 0.59$\pm$0.09
        & \textbf{20.7$\pm$1.0} & 18.8$\pm$0.8
        & 1.63$\pm$0.18 & \textbf{1.45$\pm$0.12} \\

\bottomrule
\end{tabular*}
\end{table*}

\subsection{Evaluation on In-vivo Data}\label{subsec8}

\textbf{Qualitative qMRI SR Assessment:}
Figure~\ref{fig:slice_acquired_data}a shows SR outputs for the three MuPa-ZTE protocols: fast (Acq1, 1-minute), standard (Acq5, 5-minute), and ReconDL-enhanced (Acq5$^{\text{R}}$, supervised CNN reconstruction). Visual assessment confirms consistent enhancement across all protocols: improved edge definition and reduced noise for Acq1 and Acq5 inputs, with SR outputs approaching Acq5$^{\text{R}}$ reference quality. Notably, when Acq5$^{\text{R}}$ serves as input---already enhanced by ReconDL---SWIG qMRI SR still produces subtle refinements in the SR outputs, particularly in T2 CSF quantification.

\textbf{Evaluation via wMRI Synthesis:}
Figure~\ref{fig:slice_acquired_data}b presents T1w and T2w images synthesized from $Q_l$ and $\hat{Q}_h$ using forward models \eqref{IR-GRE}, \eqref{PROPELLER} with acquisition-matched parameters. Synthesized images from $\hat{Q}_h$ demonstrate superior fidelity to acquired in-vivo references across all protocols, confirming effective performance on clinical data. Notably, wMRI synthesized from SR-enhanced Acq1 and Acq5 inputs show reduced blurring, sharper anatomical delineation, and lower noise compared to synthesis from Acq5$^{\text{R}}$.

Figure~\ref{fig:slice_acquired_data}c evaluates T2-FLAIR synthesis---a contrast never seen by the SR model. T2-FLAIR images synthesized from Acq1 and Acq5 exhibit substantial noise, partial volume artifacts, and poor fluid suppression (red arrows). In contrast, T2-FLAIR from SR-enhanced qMRI shows marked improvements for these protocols. For Acq5$^{\text{R}}$, T2-FLAIR synthesis from the input already exhibits reduced noise and adequate ventricular suppression compared to Acq1 and Acq5 inputs, consistent with its higher baseline qMRI quality. Applying SR to Acq5$^{\text{R}}$ does not yield further
improvement in T2-FLAIR synthesis fidelity in contrast to Acq1 and Acq5, where SR substantially reduces noise and improves fluid suppression.

Table~\ref{tab:metrics_inference_subjectwise} presents subject-wise metrics computed over axial slices 40--140 and reported as mean$\pm$std across slices. Synthesized wMRI from $\hat{Q}_h$ consistently achieved better mean SSIM, PSNR, and HFEN than synthesis from $Q_l$ for T1w and T2w across all protocols and subjects.

For representative Subject~1, T1w improvements were larger for Acq1 and Acq5
than for Acq5$^{\mathrm{R}}$: SSIM increased by 11.0\%, 10.8\%, and 9.4\%,
HFEN decreased by 53.3\%, 53.4\%, and 30.4\%, and PSNR increased by 2.2, 1.8,
and 1.5\,dB, respectively. This is consistent with the stronger baseline T1w
quality of Acq5$^{\mathrm{R}}$ ($Q_l$ SSIM 0.85 versus 0.82--0.83), which
leaves less headroom for improvement. The same ordering held in Subjects~2
and~3.

For T2w, PSNR increased by 5.2\,dB for Acq1, 4.7\,dB for Acq5, and 4.8\,dB
for Acq5$^{\mathrm{R}}$, and HFEN decreased by 67.4\%, 74.3\%, and 72.5\%.
The largest relative SSIM gain occurred for Acq5$^{\mathrm{R}}$ (37.7\% versus
13.0\% and 17.6\%), reflecting its substantially lower baseline T2w SSIM
(0.61 versus 0.74--0.77).

For T2-FLAIR, which was not used as guidance, Acq1 and Acq5 improved across
all three metrics. For Subject~1, SSIM increased by 8.7\% and 10.4\%, HFEN
decreased by 9.2\% and 21.7\%, and PSNR increased from 19.3 to 22.0\,dB and
from 20.8 to 22.1\,dB, respectively. For Acq5$^{\mathrm{R}}$, HFEN improved by
10.8\% while SSIM decreased by 11.8\% (0.68 to 0.60) and PSNR by 1.9\,dB
(21.0 to 19.1\,dB). This divergence was reproduced in all three subjects and
is therefore systematic rather than subject-specific.

\section{Discussion}\label{sec5}
\subsection{Principal Findings}\label{subsec9}
Physics-informed, self-supervised learning achieved effective qMRI SR using routinely acquired clinical weighted images without requiring HR qMRI for training. Ablation studies established: (1) contrast-specific enhancement (T1w was observed to mostly improve T1 maps, and T2w improves T2 maps); (2) dual-guide configuration optimally enhances all parameters; (3) LR qMRI input is essential for quantitative accuracy. Validation on independently acquired MuPa-ZTE data confirmed: (4) cross-sequence generalizability (5) improved synthesis of a contrast never used during training (T2-FLAIR), demonstrating recovery of true quantitative tissue properties. Practically, qMRI SR enables synthesis of additional
contrasts beyond those used as guides, as demonstrated by the
T2-FLAIR synthesis experiments, where synthesis quality from
$\hat{Q}_h$ improved relative to synthesis from fast
$Q_l$.

\subsection{Contrast-Specific Enhancement and LR qMRI Necessity}\label{subsec10}
T1w images provide strong spatial information about T1 distribution but weak
T2 information, and vice versa for T2w images. Model $M_{12}^{-+}$, which
excluded $Q_l$ at inference, exhibited quantitative errors---reasonable white
matter T1 recovery but failed gray matter quantification. This demonstrates
wMRI alone cannot reliably recover absolute quantitative values in all
regions. The same behavior was observed for the adapted Qiu et al.\
variant $M_{12}^{-+,Qiu}$, in which $L_h$ is replaced by a Sobel-edge guide
loss. This variant did not outperform $M_{12}^{-+}$ and yielded higher errors
for most metric--map combinations. The two guide losses constrain the
prediction differently: the Sobel-edge loss acts on high-gradient locations
and therefore leaves homogeneous tissue interiors largely unconstrained,
whereas the correlation ratio $L_h$ is evaluated over all brain-masked voxels
and additionally accounts for global intensity scaling between $W$ and
$\hat{W}$. Neither formulation constrains absolute parameter values, however,
and the comparison indicates that the limiting factor is the absence of the
subject-specific $Q_l$ anchor rather than the choice of guide loss. Without
$Q_l$, the network lacks a subject-specific $Q_l$ anchor and must infer qMRI
values solely from subject's HR wMRI and priors learned from other training
subjects' $C_l$, which can produce structurally plausible but quantitatively
inaccurate predictions.

Furthermore, model $M_{12}^{--}$ produces meaningless  $\hat{Q}$
across all regions. This demonstrates that direct wMRI-to-qMRI mapping fails when only two weighted images are available---a realistic constraint in clinical settings. Together, $M_{12}^{-+}$ and $M_{12}^{--}$ provide concrete evidence that LR qMRI is essential during both training and inference, and
highlight a fundamental advantage of SR over direct wMRI-to-qMRI
estimation~\cite{bib_b5,bib_b6,bib_b7}: by preserving acquired LR qMRI as a
subject-specific quantitative reference, SR outputs remain traceable to an actual measurement, whereas methods that bypass qMRI acquisition entirely produce maps that cannot be validated against any acquired quantitative data.

The tissue-specific MAE results (Fig.~\ref{fig:boxplots})
confirm that $M_{12}^{++}$ reduces MAE across all tissue classes and parametric maps relative to the LR baseline. This indicates that no systematic bias is introduced in any region, demonstrating that SWIG qMRI SR preserves absolute quantitative accuracy rather than merely improving structural appearance. The most pronounced improvements occur in CSF, where partial volume effects in LR maps are most severe --- precisely the regime where accurate relaxometry is clinically most challenging. As discussed above, the quantitative degradation of $M_{12}^{-+}$ in these regions, despite competitive SSIM values, further underscores the necessity of the LR qMRI anchor.

\subsection{Architectural Choice and Framework Generalizability}\label{subsec11}

The primary contribution of this work is the physics-informed
self-supervised optimization framework rather than the specific
network architecture. Since SWIG qMRI SR is formulated as a MAP
inference problem (\eqref{eq:weighted_loss}), the proposed objective is architecture-agnostic and can be optimized using any
differentiable parametric model. This is confirmed empirically: the U-Net and ResNet backbones produced comparable results under the identical loss formulation. The ResNet was retained as the primary model throughout all
experiments; the U-Net was added specifically as an architectural
sensitivity test, not as a replacement candidate, confirming that
backbone selection is a secondary design decision. Direct comparison with fully supervised SR methods such as GAN- or diffusion-based approaches was intentionally excluded, since the
proposed framework is specifically designed to avoid dependence on high-resolution qMRI training targets. Nevertheless, such architectures could in principle
serve as alternative backbones within the proposed self-supervised objective, and exploring more expressive architectures remains a promising direction for future work.

Moreover, the balance between the two loss terms in
\eqref{eq:weighted_loss} is critical: the sensitivity analysis of
$\alpha$ identifies a range ($\alpha \in [10^{1}, 10^{4}]$) in
which structural guidance improves performance while quantitative
consistency with $C_l$ is maintained. Outside this range,
performance degrades: at low values ($\alpha \leq 10^{0}$), guide
contribution is negligible; beyond $\alpha = 10^{4}$, the guide
loss dominates and quantitative accuracy degrades
sharply.

\subsection{Cross-Sequence Generalizability}\label{subsec12}
SWIG qMRI SR, trained on 2D QRAPMASTER, successfully enhanced 3D MuPa-ZTE maps despite fundamental differences: $k$-space sampling trajectory (Cartesian vs. radial), acquisition timings, dimensionality, and reconstruction algorithms. SR outputs from 1-minute Acq1 exceeded 5-minute Acq5 quality (Fig.~\ref{fig:slice_acquired_data}a, Table~\ref{tab:metrics_inference_subjectwise}).

This generalizability arises from operating on relaxometry parameters---intrinsic tissue properties independent of acquisition protocol---rather than sequence-dependent signal intensities. Methods operating on $k$-space~\cite{bib_b3} or raw wMRI intensities must learn sequence-specific relationships requiring retraining for different protocols.

The clinical implications of the cross-sequence generalizability of the implementation are significant. SWIG qMRI SR enables train-once, deploy-across-sequences workflow: the key requirement is matching guide protocols (clinically standardized T1w IR-SPGR, T2w SE), while qMRI sequences can vary. Nevertheless, validation on each target sequence remains essential. This work validates one sequence (MuPa-ZTE) in three subjects. Further validation across additional sequences (MP2RAGE~\cite{bib_b28}, DESPOT~\cite{bib_b29}, MR fingerprinting~\cite{bib_b17}) and pathological populations is needed.

\subsection{Comparison with Supervised Approaches}\label{subsec13}

$M^{s}$ defines the performance ceiling under the assumption
that HR qMRI is available for training within the same experimental
setting. Relative to $M_{12}^{++}$, $M^{s}$ achieved higher SSIM,
lower HFEN, and lower tissue-specific MAE. This gap quantifies the cost of replacing
voxel-wise HR qMRI supervision with the proposed physics-informed
self-supervised objective. This cost is concentrated where the
guide loss alone cannot constrain absolute quantitative values.
Closing this remaining margin without reintroducing HR qMRI as a
training requirement is an open direction for future work.

A distinct comparison arises with Acq5$^{\text{R}}$, where
supervision targets raw k-space reconstruction rather than qMRI SR
and no independently acquired wMRI guides are available to the
reconstruction network. In this setting, SWIG qMRI SR achieved
substantial improvements when applied to Acq1 and Acq5 inputs,
with SR-enhanced outputs surpassing even Acq5$^{\text{R}}$ maps.
This superiority is evident both in visual inspection
(Fig.~\ref{fig:slice_acquired_data}a) and quantitative metrics for
synthesized images (Table~\ref{tab:metrics_inference_subjectwise}).
Specifically, T1w/T2w synthesized from SR-enhanced Acq1 and Acq5 maps
consistently outperformed wMRI synthesized from Acq5$^{\text{R}}$
across all three subjects, demonstrating
that SWIG qMRI SR recovers quantitative information beyond what
supervised reconstruction (ReconDL) alone achieves.

This indicates complementary information sources: ReconDL learns spatial priors from training data, while SWIG qMRI SR leverages structural information from independently acquired clinical wMRI. Such supervised reconstruction cannot access information in wMRI beyond its training data, particularly valuable for abnormalities never seen by supervised networks.

\subsection{Unseen T2-FLAIR Synthesis Validates Quantitative Enhancement}\label{subsec14}
T2-FLAIR synthesis---never used as a guide---provides compelling evidence of true quantitative recovery rather than feature copying. Synthesis from super-resolved maps outperformed LR inputs across Acq1 and Acq5. The successful synthesis of T2-FLAIR, which requires different parameter combinations than the guide contrasts, confirms that SR genuinely improved quantitative tissue characterization rather than merely copying structural features from guides.

For Acq5$^{\text{R}}$, T2-FLAIR synthesis from $Q_l$ showed adequate ventricular suppression but failed in narrow cortical sulci, producing bright artifacts; synthesis from $\hat{Q}_h$ showed some visual reduction of these artifacts, though residual artifacts remained visible. Notably, despite this localized visual difference, SR degraded T2-FLAIR synthesis quality overall for Acq5$^{\text{R}}$ in SSIM and PSNR (Table~\ref{tab:metrics_inference_subjectwise}), in contrast to the improvement observed for Acq1 and Acq5. Acq5$^{\text{R}}$ is the protocol where the LR qMRI input already approaches the quality of the SR outputs obtained from Acq1 and Acq5, suggesting that once the LR qMRI anchor is already of sufficiently high quality, SR may offer no further benefit, and can even degrade synthesis of an unseen contrast such as T2-FLAIR. This points to the need for explicit evaluation across a range of LR qMRI quality levels and acceleration factors to characterize the threshold beyond which SR ceases to be beneficial.

For the  protocols Acq1 and Acq5, however, these synthesis results carry direct clinical relevance: since clinicians rely predominantly on conventional weighted images for diagnosis, the ability to synthesize higher-fidelity diagnostic contrasts---including T2-FLAIR, among the most diagnostically sensitive contrasts in neuroimaging---from a 1-minute qMRI acquisition represents a practical benefit of HR qMRI in routine workflows.

Note that T2-FLAIR synthesis uses the simplified signal model in \eqref{T2-FLAIR-signal-equation}. Since $Q_l$- and $\hat{Q}_h$-derived syntheses share this model, the relative improvement from $\hat{Q}_h$ remains valid, though absolute synthesis fidelity and residual artifacts should be interpreted with this simplification in mind.

\subsection{Limitations of the Current Work}\label{subsec15}

\textbf{Training and Validation on pathological cases:} In this work, training used synthetic data from healthy volunteers and in-vivo validation was limited to three healthy volunteer, yet the primary motivation is training with qMRI from patient populations. This setup is sufficient as proof-of-concept of the proposed framework under controlled conditions. Moreover, pathological tissues may violate mono-exponential relaxation assumptions (e.g., tumors with mixed compartments), potentially causing systematic errors. Having validated the concept, the framework's independence from HR qMRI supervision opens a practical path forward: rather than acquiring HR qMRI in patients, future work can prioritize acquiring fast LR qMRI alone, which is sufficient to train the framework on clinically relevant, pathological data.
When validating that trained model, it is relevant to evaluate potential deviations from the mono-exponential signal models employed, for example in edema, necrosis, or multi-compartment effects.

\textbf{Signal Model Simplifications:}
The framework relies on two physical models, both of which use
simplified signal equations. The first model generates LR qMRI from the
resolved HR qMRI and enforces consistency with the acquired quantitative maps.
These forward models neglect practical effects such as variable-flip-angle
refocusing in fast spin-echo acquisitions~\cite{bib_b30}, slice-profile
imperfections, and \(B_1^+\) inhomogeneities. Consequently, biases already
present in the acquired LR qMRI maps may be propagated through the
data-consistency constraint rather than corrected by the proposed framework.
More sophisticated signal representations, such as the Extended Phase Graph
formalism~\cite{bib_b31}, could model multi-echo acquisitions and imperfect
refocusing more accurately.

The second model predicts the weighted image from the resolved HR qMRI.
In this model, \(B_1^+\) inhomogeneity is particularly relevant because the
predicted signal depends explicitly on the nominal flip angle. Spatial
variations in the effective flip angle may therefore produce structured
discrepancies between the acquired weighted image, \(W\), and its
forward-model prediction, \(\hat{W}\). The correlation-ratio guide loss is
insensitive to global intensity scaling and offset, but it does not eliminate
spatially varying signal biases. Such discrepancies could therefore affect the
correlation-guided component of training and potentially bias the recovered HR
qMRI maps.

Slice-profile and other point-spread-function imperfections introduce a
related limitation. The framework assumes an idealized relationship between
the HR and LR images, whereas non-ideal slice excitation and
acquisition-dependent blurring may alter the effective resolution and edge
sharpness of both the quantitative and weighted images. These effects are not
fully represented by the \(k\)-space truncation used to generate the synthetic
LR data and may therefore cause the synthetic experiments to provide an
optimistic estimate of performance for acquisitions with more pronounced
imperfections.

The successful application of the method to acquired MuPa-ZTE data,
which is inherently subject to scanner-related field and acquisition
imperfections, provides indirect evidence that the framework can operate in
their presence. However, this does not constitute a controlled sensitivity
analysis or establish that these effects have a negligible influence on
quantitative accuracy. Direct evaluation would require, for example,
simulations with controlled \(B_1^+\) variations and slice-profile
distortions, or phantom and volunteer experiments incorporating measured
\(B_1^+\) maps. We therefore identify systematic evaluation of these model
mismatches as an important direction for future work.

Relatedly, absolute quantitative accuracy against a standardized
ISMRM/NIST phantom was not established in this work; the held-out synthetic
data and independently acquired in-vivo data used here support relative
validation but not absolute accuracy across the full physiological T1/T2
range, and phantom-based benchmarking remains necessary before clinical
translation.

\textbf{Acceleration Factor and LR qMRI Quality:} The current
work evaluates a single acceleration level: Acq1 corresponds to
approximately 5-fold acceleration relative to the full-duration
protocol. The ablation experiments partially characterize the
sensitivity to LR qMRI - $M_{12}^{--}$ and $M_{12}^{-+}$
represent the extreme case of no quantitative anchor, confirming that its absence degrades performance, while $M_{12}^{++}$ demonstrates that LR qMRI provides a sufficient anchor. However,
the intermediate regime where LR qMRI is present but of
progressively lower quality due to increased acceleration has not
been systematically evaluated. Determining the lower bound on usable acceleration, below which the LR qMRI anchor becomes insufficient for SR, remains an open question that depends on sequence-specific SNR characteristics and the structural detail
available in the selected wMRI guides.

Additional open questions include cross-field-strength
generalizability---as deployment at a different field strength requires dedicated evaluation---and robustness to motion artifacts. The synthetic training pipeline uses artifact-free data and the framework has no explicit mechanism for handling motion corruption. In the
in-vivo validation, rigid-body registration was applied to align
acquired qMRI and guide images, mitigating gross misalignment, but intra-acquisition motion and motion-induced artifacts in the qMRI maps themselves were not addressed. As with any learning-based framework, incorporating motion-corrected acquisitions will be essential for robust clinical deployment. Moreover  sequence- or tissue-specific tuning of weighting factors of loss function, for instance in acquisitions with substantially different SNR regimes --- remains an open direction.

\subsection{Outlook}\label{subsec16}
The present work demonstrates a practical step towards integrating rapid qMRI into routine protocols. By adding a rapid qMRI acquisition and combining it with existing clinical wMRI scans, we showed it is possible to do qMRI SR. This highlights a compelling use-case: a standard MRI exam could include a brief qMRI sequence (e.g., 1 minute MuPa-ZTE) and leverage the standard images to obtain HR qMRI post hoc. The overall scan time overhead is minimal, yet the  benefits are significant---both in terms of richer quantitative information and the ability to derive missing image contrasts on demand.

\section{Conclusion}\label{sec6}

We introduced SWIG qMRI SR, a physics-informed, self-supervised DL framework for qMRI SR by leveraging conventional clinical wMRIs. Our approach avoids the need for HR qMRI during training, instead relying on the physical relationship between qMRI and wMRI to guide learning. Through extensive experiments on both synthetic and acquired data, we demonstrated that SWIG qMRI SR enhances the resolution and fidelity of qMRI, while also enabling the synthesis of realistic wMRI. The framework generalized well to independently acquired data, indicating robustness to a different qMRI pulse sequence. Furthermore, the ability to synthesize missing MRI contrasts from the SR outputs supports integration of SWIG qMRI SR into fast qMRI workflows, potentially reducing overall scan time and improving clinical applicability. Overall, SWIG qMRI SR represents a promising approach to making  qMRI more accessible and compatible with routine clinical practice.

\backmatter

\bmhead{Acknowledgements}

This work was conducted within the ``Trustworthy AI for MRI'' ICAI lab within the project ROBUST, funded by the Dutch Research Council, GE HealthCare, and the Dutch Ministry of Economic Affairs and Climate Policy.

\section*{Declarations}

\textbf{Funding} This work was conducted within the ``Trustworthy AI for MRI'' ICAI lab within the project ROBUST, funded by the Dutch Research Council, GE HealthCare, and the Dutch Ministry of Economic Affairs and Climate Policy.

\textbf{Competing interests} Florian Wiesinger is employed by GE HealthCare. The remaining authors have no relevant financial or non-financial interests to disclose.

\textbf{Ethics approval} This study was performed in line with the principles of the Declaration of Helsinki. All data acquisition protocols were approved by the institutional review board of Erasmus MC.

\textbf{Consent to participate} Informed consent was obtained from all individual participants included in the study.

\textbf{Consent to publish} All participants provided written informed consent for publication of their anonymized imaging data.

\textbf{Data availability} The datasets generated and analysed during the current study are available from the corresponding author on reasonable request.

\textbf{Code availability} The code required to reproduce the results reported in this article will be made available upon acceptance. The trained models are available from the corresponding author on reasonable request.

\textbf{Author contributions} Samadifardheris: conceptualization, methodology, software, validation, formal analysis, investigation, data curation, writing---original draft, writing---review and editing, visualization. Poot: conceptualization, methodology, validation, formal analysis, investigation, resources, data curation, writing---review and editing, visualization, supervision, project administration, funding acquisition. Wiesinger: conceptualization, methodology, investigation, writing---review and editing, supervision. Klein: conceptualization, methodology, writing---review and editing, supervision, funding acquisition. Hernandez-Tamames: conceptualization, methodology, investigation, resources, writing---review and editing, supervision, funding acquisition. All authors read and approved the final manuscript.


\begin{thebibliography}{00}

\bibitem{bib_b1} Cashmore MT, McCann AJ, Wastling SJ, McGrath C, Thornton J, Hall MG (2021) Clinical quantitative MRI and the need for metrology. Br J Radiol 94:20201215

\bibitem{bib_b2} Whittall KP, MacKay AL (1989) Quantitative interpretation of NMR relaxation data. J Magn Reson 84:134--152

\bibitem{bib_b3} Liu F, Feng L, Kijowski R (2019) MANTIS: model-augmented neural network with incoherent $k$-space sampling for efficient MR parameter mapping. Magn Reson Med 82:174--188

\bibitem{bib_b4} Chaudhari AS et al (2018) Super-resolution musculoskeletal MRI using deep learning. Magn Reson Med 80:2139--2154

\bibitem{bib_b5} Qiu S, Christodoulou AG, Xie Y, Li D (2022) Hybrid supervised and self-supervised deep learning for quantitative mapping from weighted images using LR labels. In: Proceedings of the annual meeting, International Society for Magnetic Resonance in Medicine, London, p 2609

\bibitem{bib_b6} Qiu S et al (2022) Multiparametric mapping in the brain from conventional contrast-weighted images using deep learning. Magn Reson Med 87:488--495

\bibitem{bib_b7} Moya-S\'aez E, Pe\~na-Nogales \'O, de Luis-Garc\'ia R, Alberola-L\'opez C (2021) A deep learning approach for synthetic MRI based on two routine sequences and training with synthetic data. Comput Methods Programs Biomed 210:106371

\bibitem{bib_b8} Atal\i k A, Chopra S, Sodickson DK (2025) A trust-guided approach to MR image reconstruction with side information. IEEE Trans Med Imaging. doi:10.1109/TMI.2025.3594363

\bibitem{bib_b9} Roche A, Malandain G, Ayache N (2000) Unifying maximum likelihood approaches in medical image registration. Int J Imaging Syst Technol 11:71--80

\bibitem{bib_b10} Warntjes JBM, Leinhard OD, West J, Lundberg P (2008) Rapid magnetic resonance quantification on the brain: optimization for clinical usage. Magn Reson Med 60:320--329

\bibitem{bib_b11} Tanenbaum LN et al (2017) Synthetic MRI for clinical neuroimaging: results of the magnetic resonance image compilation (MAGiC) prospective, multicenter, multireader trial. AJNR Am J Neuroradiol 38:1103--1110

\bibitem{bib_b12} Isensee F et al (2019) Automated brain extraction of multi-sequence MRI using artificial neural networks. Hum Brain Mapp 40:5065--5077

\bibitem{bib_b13} Zhang Y, Brady M, Smith S (2001) Segmentation of brain MR images through a hidden Markov random field model and the expectation-maximization algorithm. IEEE Trans Med Imaging 20:45--57

\bibitem{bib_b14} Wiesinger F, McKinnon G, Kaushik S (2021) 3D silent parameter mapping: further refinements and quantitative assessment. In: Proceedings of the annual meeting, International Society for Magnetic Resonance in Medicine, virtual, p 1828

\bibitem{bib_b15} Ljungberg E, Wood TC, Solana AB, Barker GJ, Wiesinger F (2022) Silent motion-corrected ZTE for quantitative T1 and T2 mapping. In: Proceedings of the annual meeting, International Society for Magnetic Resonance in Medicine, London, p 3287

\bibitem{bib_b16} Pipe JG, Menon P (1999) Sampling density compensation in MRI: rationale and an iterative numerical solution. Magn Reson Med 41:179--186

\bibitem{bib_b17} Ma D et al (2013) Magnetic resonance fingerprinting. Nature 495:187--192

\bibitem{bib_b18} Wiesinger F et al (2020) PSST...parameter mapping swift and silent. In: Proceedings of the annual meeting, International Society for Magnetic Resonance in Medicine, virtual, p 0882

\bibitem{bib_b19} Mandava S, Carl M, Wiesinger F, Fung M, Lebel RM (2024) Deep learning-based chemical shift artifact reduction in zero echo time (ZTE) MRI. In: Proceedings of the annual meeting, International Society for Magnetic Resonance in Medicine, Singapore, p 4428

\bibitem{bib_b20} Lebel RM (2020) Performance characterization of a novel deep learning-based MR image reconstruction pipeline. White paper, GE Healthcare, Calgary

\bibitem{bib_b21} Boucneau T et al (2021) AZTEK: adaptive zero TE $k$-space trajectories. Magn Reson Med 85:926--935

\bibitem{bib_b22} Klein S, Staring M, Murphy K, Viergever MA, Pluim JPW (2010) elastix: a toolbox for intensity-based medical image registration. IEEE Trans Med Imaging 29:196--205

\bibitem{bib_b23} Shamonin DP, Bron EE, Lelieveldt BPF, Smits M, Klein S, Staring M (2014) Fast parallel image registration on CPU and GPU for diagnostic classification of Alzheimer's disease. Front Neuroinform 7:50

\bibitem{bib_b24} He K, Zhang X, Ren S, Sun J (2016) Deep residual learning for image recognition. In: Proceedings of the IEEE conference on computer vision and pattern recognition, Las Vegas, pp 770--778

\bibitem{bib_b25} Ronneberger O, Fischer P, Brox T (2015) U-Net: convolutional networks for biomedical image segmentation. In: Navab N, Hornegger J, Wells WM, Frangi AF (eds) Medical image computing and computer-assisted intervention -- MICCAI 2015. Springer, Cham, pp 234--241

\bibitem{bib_b26} Wang Z, Bovik AC, Sheikh HR, Simoncelli EP (2004) Image quality assessment: from error visibility to structural similarity. IEEE Trans Image Process 13:600--612

\bibitem{bib_b27} Ravishankar S, Bresler Y (2011) MR image reconstruction from highly undersampled $k$-space data by dictionary learning. IEEE Trans Med Imaging 30:1028--1041

\bibitem{bib_b28} Marques JP et al (2010) MP2RAGE, a self bias-field corrected sequence for improved segmentation and T1-mapping at high field. Neuroimage 49:1271--1281

\bibitem{bib_b29} Deoni SC, Rutt BK, Peters TM (2003) Rapid combined T1 and T2 mapping using gradient recalled acquisition in the steady state. Magn Reson Med 49:515--526

\bibitem{bib_b30} Busse RF, Hariharan H, Vu A, Brittain JH (2006) Fast spin echo sequences with very long echo trains: design of variable refocusing flip angle schedules and generation of clinical T2 contrast. Magn Reson Med 55:1030--1037

\bibitem{bib_b31} Weigel M (2015) Extended phase graphs: dephasing, RF pulses, and echoes -- pure and simple. J Magn Reson Imaging 41:266--295

\end{thebibliography}
\end{document}